\documentclass[11pt]{article}

\usepackage[final]{acl}
\usepackage{times}
\usepackage{latexsym}
\usepackage[T1]{fontenc}
\usepackage[utf8]{inputenc}
\usepackage{microtype}
\usepackage{inconsolata}
\usepackage{graphicx}
\usepackage{amsmath}
\usepackage{amsfonts}
\usepackage{multirow}
\usepackage{booktabs}
\usepackage{setspace}
\usepackage{enumitem}
\usepackage{cleveref}
\usepackage{subcaption}
\usepackage{graphicx}
\usepackage{algorithm}
\usepackage{algpseudocode}
\usepackage[utf8]{inputenc}
\usepackage{colortbl,xcolor}
\usepackage{tabularx}
\usepackage{stfloats}
\crefformat{section}{§#2#1#3}
\usepackage{textcomp}

\title{MASPO: Unifying Gradient Utilization, Probability Mass, and Signal Reliability for Robust and Sample-Efficient LLM Reasoning}

\author{
    Xiaoliang Fu\textsuperscript{1,2,*}, 
    Jiaye Lin\textsuperscript{1,*,$\dagger$}, 
    Yangyi Fang\textsuperscript{1,3,*}, 
    Binbin Zheng\textsuperscript{1,4}, 
    Chaowen Hu\textsuperscript{1}, \\
    \textbf{Zekai Shao}\textsuperscript{2}, 
    \textbf{Cong Qin}\textsuperscript{1,5}, 
    \textbf{Lu Pan}\textsuperscript{1}, 
    \textbf{Ke Zeng}\textsuperscript{1}, 
    \textbf{Xunliang Cai}\textsuperscript{1} \\
    \textsuperscript{1}Meituan \quad
    \textsuperscript{2}Fudan University \quad
    \textsuperscript{3}Tsinghua University \\
    \textsuperscript{4}University of Science and Technology of China \quad
    \textsuperscript{5}Peking University \\
    \texttt{\{fuxiaoliang04, linjiaye\}@meituan.com}
}

\definecolor{mistyblue}{RGB}{230, 237, 245}
\definecolor{almond}{RGB}{250, 241, 230}
\definecolor{ourshighlight}{gray}{0.95}
\definecolor{bestcolor}{RGB}{55, 85, 140}
\definecolor{secondcolor}{RGB}{0, 100, 80}
\definecolor{biascolor}{RGB}{192, 0, 0}
\definecolor{improvecolor}{RGB}{45, 60, 145}

\newcommand{\best}[1]{\textbf{#1}}
\newcommand{\second}[1]{\underline{#1}}

\begin{document}

\maketitle

\begingroup
  \renewcommand\thefootnote{}
  \footnotetext{
    \textsuperscript{*}\ Equal contribution. \quad 
    \textsuperscript{$\dagger$}\ Corresponding author.
  }
\endgroup

\begin{abstract}
Existing Reinforcement Learning with Verifiable Rewards (RLVR) algorithms, such as GRPO, rely on rigid, uniform, and symmetric trust region mechanisms that are fundamentally misaligned with the complex optimization dynamics of Large Language Models (LLMs). In this paper, we identify three critical challenges in these methods: (1) \textit{inefficient gradient utilization} caused by the binary cutoff of hard clipping, (2) \textit{insensitive probability mass} arising from uniform ratio constraints that ignore the token distribution, and (3) \textit{asymmetric signal reliability} stemming from the disparate credit assignment ambiguity between positive and negative samples. To bridge these gaps, we propose \textbf{M}ass-\textbf{A}daptive \textbf{S}oft \textbf{P}olicy \textbf{O}ptimization (\textbf{MASPO}), a unified framework designed to harmonize these three dimensions. MASPO integrates a differentiable soft Gaussian gating to maximize gradient utility, a mass-adaptive limiter to balance exploration across the probability spectrum, and an asymmetric risk controller to align update magnitudes with signal confidence. Extensive evaluations demonstrate that MASPO serves as a robust, all-in-one RLVR solution, significantly outperforming baselines. Our code is available at: \href{https://github.com/FlyTune/MASPO-RL}{https://github.com/FlyTune/MASPO-RL}.
\end{abstract}

\section{Introduction}

Reinforcement Learning with Verifiable Rewards (RLVR) has evolved into the cornerstone of reasoning optimization in Large Language Models (LLMs)~\cite{lightman2023let, shao2024deepseekmath}. While algorithms like GRPO~\cite{shao2024deepseekmath} eliminate the need for specific critic models, they rely on a ``hard clipping'' mechanism inherited from PPO~\cite{schulman2017proximal}. We rethink that it is a fragmented solution, failing to address the complex interplay between heavy-tailed vocabulary distributions and sparse reward signals.

\begin{figure*}
\centering
\includegraphics[width=1.0\linewidth]{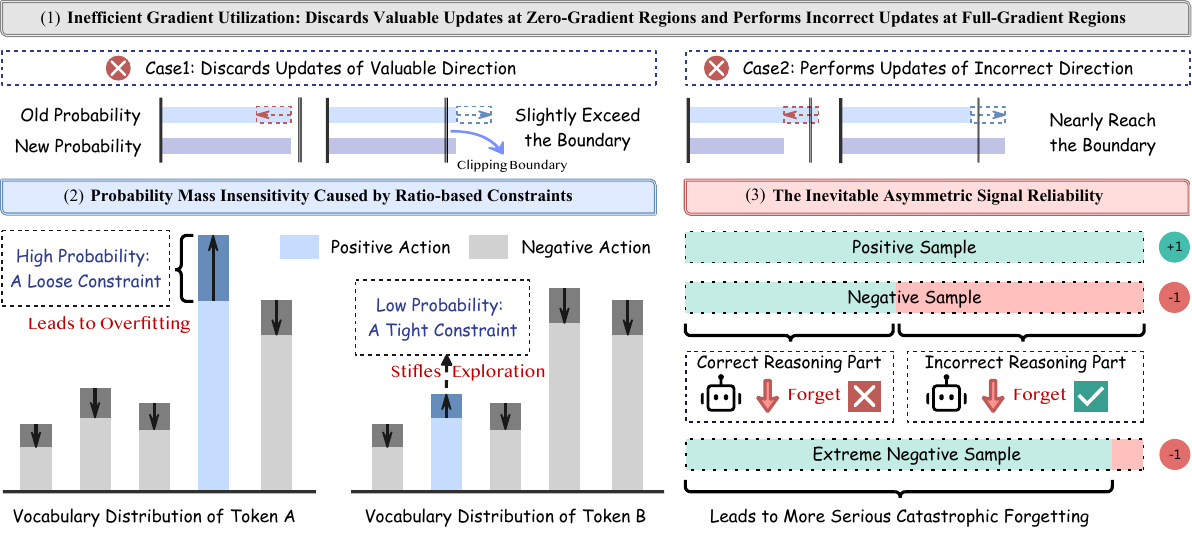}
\vspace{-1em}
\caption{MASPO resolves three core limitations of GRPO: (1) \textit{inefficient gradient utilization} caused by hard boundaries that indiscriminately discard valid gradients; (2) \textit{insensitive probability mass}, where uniform clipping overlooks the head/tail mass disparity; and (3) \textit{asymmetric signal reliability}, which neglects the inherent noise differences between verified positive and ambiguous negative signals.}
\vspace{-1.5em}
\label{fig:intro}
\end{figure*}

As illustrated in Figure \ref{fig:intro}, current paradigms suffer from three distinct misalignments: 
(1) \textit{Inefficient Gradient Utilization}. Hard clipping mechanism imposes a rigid binary cutoff that indiscriminately discards valuable reinforcement signals from successful explorations exceeding the predefined boundary, while simultaneously failing to provide corrective gradients for severe errors that are within the boundary, thus diminishing the effective utilization of generated samples.
(2) \textit{Insensitive Probability Mass}. A uniform clipping range (e.g., $[1 - \varepsilon, 1 + \varepsilon]$) neglects the token's absolute probability. It imposes overly loose constraints on high-probability tokens (risking policy collapse) while excessively restricting low-probability tokens in the long tail (hindering exploration).
(3) \textit{Asymmetric Signal Reliability}. Positive rewards verify correct reasoning, whereas negative rewards suffer from credit assignment ambiguity. Treating their advantage estimates symmetrically overlooks their disparate Signal-to-Noise Ratios (SNR).

Aiming to address these challenges holistically, we propose \textbf{MASPO}, a \textbf{M}ass-\textbf{A}daptive \textbf{S}oft \textbf{P}olicy \textbf{O}ptimization framework, which designed to synthesize trust region management into a single, cohesive objective. By replacing the rigid box constraint with a soft Gaussian gating mechanism, MASPO resolves the gradient inefficiency. Crucially, it synergizes two adaptive components: a mass-adaptive limiter that expands the exploration budget for low-probability tokens to address mass insensitivity and an asymmetric risk controller that modulates update magnitudes based on signal confidence to resolve reliability asymmetry. Our contributions in this paper are summarized as follows:
\begin{itemize}[leftmargin=0.5cm]
\item \textbf{Unified Perspective:} We systematically identify inherent challenges in current trust region paradigms and propose a holistic perspective that aligns RLVR optimization by addressing three fundamental misalignments.
\item \textbf{All-in-One Framework:} We propose a comprehensive solution, which unifies a soft Gaussian gating, a mass-adaptive limiter, and an asymmetric risk controller into a single framework.
\item \textbf{Superior Performance:} Comprehensive evaluations on diverse benchmarks demonstrate that MASPO achieves superior sample efficiency and reasoning performance. Further experiments confirm its robustness and effectiveness in stabilizing long-chain reasoning.
\end{itemize}

\section{Related Works}

\subsection{RLVR in LLMs}
RLVR enhances LLM reasoning using rule-based signals, unlike preference-based RLHF~\cite{ouyang2022training}. While PPO~\cite{schulman2017proximal} is a standard approach, its critic dependence limits scalability. Recent work favors outcome-based supervision \cite{uesato2022solving} via group-based optimization. GRPO~\cite{shao2024deepseekmath} eliminates the critic by normalizing rewards within groups. However, it struggle with sparse rewards and high variance, necessitating stable variants~\cite{bai2026ttvs}. Beyond optimization, recent work also seeks to enhance reasoning quality through improved model architectures~\cite{dong2026neureasonerexplainablecontrollableunified} and inference-time strategies~\cite{jiang2026foeforesterrorsmakes}.

\subsection{Optimization Directions for GRPO}
Recent studies have attempted to address the limitations of the standard GRPO framework. We categorize these efforts according to the three critical misalignments identified in this paper:

\paragraph{Addressing Inefficient Gradient Utilization.}
Standard hard clipping discards valuable gradients from exploratory tokens that exceed the trust region boundary. To mitigate this \textit{Inefficient Gradient Utilization}, soft clipping strategies have been proposed to retain partial gradients. Approaches like CISPO~\cite{chen2025minimax}, GPPO~\cite{su2025klear}, CE-GPPO~\cite{su2025gppo} and ASPO~\cite{wang2025aspo} introduce smooth piecewise decay functions. SAPO~\cite{gao2025soft} goes a step further by removing boundaries entirely, employing a global gating mechanism.

\paragraph{Addressing Insensitive Probability Mass.}
Uniform Importance Sampling (IS) \cite{precup2000eligibility} constraints overly restrict exploration in low-probability regions, reflecting the issue of \textit{insensitive probability mass}. While entropy regularization~\cite{williams1991function, o2016combining, eysenbach2021maximum} encourages distribution uniformity, it indiscriminately encourages all~\cite{he2025skywork, huang2025low}. Clip Higher~\cite{yu2025dapo} relaxes upper bounds globally but ignores token-specific probabilities. DAC~\cite{yang2025dcpo}, a more aligned method, adapts bounds based on policy probabilities.

\paragraph{Addressing Asymmetric Signal Reliability.}
Positive samples are crucial for LLM reasoning but are frequently outweighed by negative ones in GRPO~\cite{xiong2025minimalist}, highlighting the issue of \textit{Asymmetric Signal Reliability}. Methods like Advantage Reweighting~\cite{yang2025not} and BAPO~\cite{xi2025bapo} boost positive sample impact by modifying advantages or clipping bounds. Other approaches achieve implicit rebalancing via relaxed bounds~\cite{yu2025dapo}, asymmetric soft clipping~\cite{su2025gppo}, or adding entropy terms to the advantage~\cite{cheng2025reasoning}.

\section{Preliminary}

\paragraph{Problem Formulation.}
We consider the task of optimizing an LLM for reasoning tasks within the framework of Reinforcement Learning with Verifiable Rewards (RLVR). Let $\mathcal{D} = \{q\}$ represent a dataset of queries, where the policy $\pi_\theta$ generates a response $o$ for a given input $q$. The correctness of the generated response is evaluated by a deterministic, rule-based reward function $r(q, o)$. Adopting the Group Relative Policy Optimization (GRPO)~\cite{shao2024deepseekmath} paradigm, for each query $q$, we sample a group of $G$ outputs $\{o_1, o_2, \dots, o_G\}$ from the current old policy $\pi_{\theta_{\text{old}}}$. The advantage estimate $\hat{A}_i$ for the $i$-th output is derived by standardizing the rewards within the group: $\hat{A}_i = (r_i - \mu_r) / \sigma_r$, where $\mu_r$ and $\sigma_r$ denote the mean and standard deviation of the rewards within the sampled group, respectively.

\paragraph{Gradient Estimation and Utilization Efficiency.}
To analyze the \textit{Inefficient Gradient Utilization} caused by rigid constraints, we formulate the policy gradient \cite{williams1992simple} in a generalized form. Let $\rho_{i,t}(\theta) = \frac{\pi_\theta(o_{i,t}|q,o_{i,<t})}{\pi_{\theta_{\text{old}}}(o_{i,t}|q,o_{i,<t})}$ be the importance sampling ratio. The objective can be unified as:
\begin{flalign}
&\nabla_\theta\mathcal{J}(\theta) = \mathbb{E}_{q \sim \mathcal{D}, \{o_i\}_{i=1}^G \sim \pi_{\theta_{\text{old}}}(\cdot|q)} \frac{1}{\sum_{i=1}^G |o_i|}\sum_{i=1}^G & \nonumber \\
&\sum_{t=1}^{|o_i|} \mathcal{F}_{i,t}(\rho) \rho_{i,t}(\theta) \hat{A}_{i} \nabla_\theta \log \pi_\theta(o_{i,t}|q,o_{i,<t}),
\end{flalign}
where $\mathcal{F}_{i,t}(\rho)$ is the gradient weighting function. Standard methods like GRPO employ a binary weighting (hard clipping) where $\mathcal{F}(\rho) = \mathbb{I}(|\rho-1| \le \varepsilon)$, which zeros out gradients outside the trust region. Attempts to improve utilization, such as SAPO~\cite{gao2025soft}, propose a continuous decay function to smooth this boundary:
\begin{flalign}
& \mathcal{F}_{i,t}^{\text{SAPO}}(\rho) = 4 p_{i,t}(\theta)(1- p_{i,t}(\theta)) \nonumber \\
& p_{i,t}(\theta) = \sigma\left(\tau_{i,t}(\rho_{i,t}(\theta)-1)\right),
\end{flalign}
where $\sigma(x)=1/(1+e^{-x})$ is the sigmoid function.

\paragraph{Probability Mass Sensitivity in Constraints.}
To address \textit{insensitive probability mass}, prior works have attempted to adjust constraints based on distribution characteristics. Entropy regularization ~\cite{o2016combining} adds a generic penalty $\beta_\text{E} \nabla_\theta \mathcal{H}(\pi_\theta)$. Clip Higher~\cite{yu2025dapo} proposes decoupled boundaries: $\varepsilon_\text{low}$ and $\varepsilon_\text{high}$. DAC ~\cite{yang2025dcpo} provides a more rigorous formulation by defining the boundaries as functions of the old policy probability $\pi_{\theta_{\text{old}}}$, acknowledging that low-probability tokens require wider trust regions:
\begin{align}
& 0.5+\frac{1}{2}\sqrt{\max(1-\frac{4\varepsilon_\text{low}}{\pi_{\theta_\text{old}}},0)} \leq \rho_{i,t}(\theta) & \nonumber \\
& \qquad \leq 0.5+\frac{1}{2}\sqrt{1+\frac{4\varepsilon_\text{high}}{\pi_{\theta_\text{old}}}}.
\end{align}
While theoretically grounded, this approach retains the rigid ``box constraint'' structure rather than a continuous adaptation.

\paragraph{Asymmetric Signal Handling.}
To address \textit{Asymmetric Signal Reliability}, several methods skew contributions toward positive samples. Advantage Reweighting ~\cite{yang2025not} statistically increases positive contributions: $\hat{A}_{i, t}=\left[\alpha_\text{A} \cdot \pi_{\theta}\left(o_{i, t}\right)+(1-\alpha_\text{A})\right] \cdot \hat{A}_{i, t}^\text{old}$. Entropy Advantage ~\cite{cheng2025reasoning} adds entropy to the advantage term. BAPO ~\cite{xi2025bapo} explicitly dynamically adjusts the clipping bounds to ensure positive sample contribution volume:
\begin{flalign}
\frac{\left|
\sum_{\substack{\hat{A}}} \pi_\theta \left[ \min\left(\rho_t  \hat{A}, \text{clip}(\rho_t, c_\text{low}, c_{\text{high}}) \hat{A} \right) \right] \right|}{\left|
\sum_{\substack{\hat{A} > 0}} \pi_\theta \left[ \min\left(\rho_t  \hat{A}, \text{clip}(\rho_t, 0, c_{\text{high}}) \hat{A} \right) \right] \right|} \geq \rho_0,
\end{flalign}
where $\rho_0$ is a threshold parameter, and $\rho_t$ is IS ratio.

\begin{figure*}[htbp]
\centering
 \includegraphics[width=1.0\linewidth]{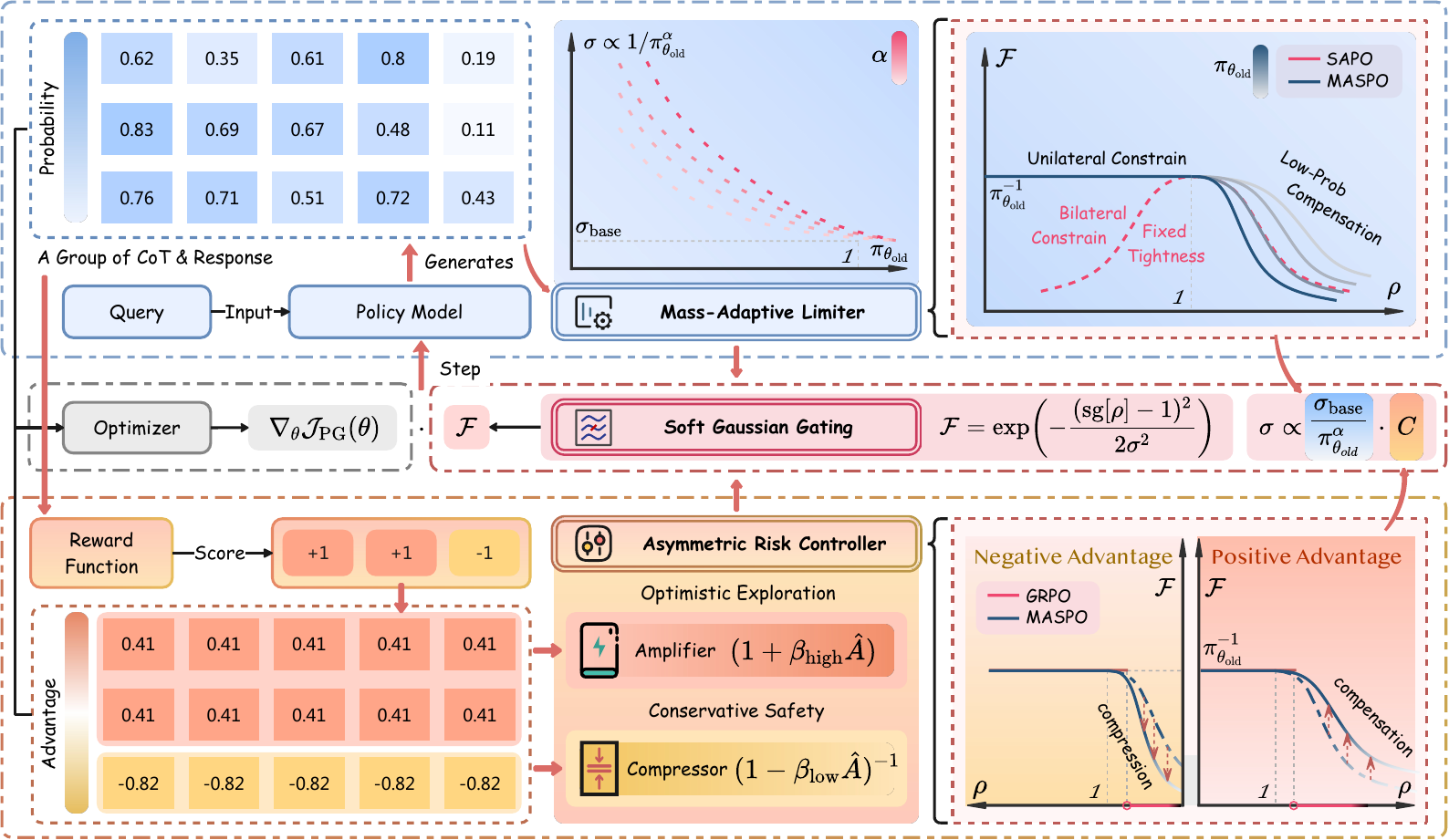}
\caption{Overview of the \textbf{MASPO} framework. The architecture integrates a \textbf{Mass-Adaptive Limiter} to scale constraints inversely with token probability and an \textbf{Asymmetric Risk Controller} to modulate update magnitude based on advantage signals, unified via a differentiable \textbf{Soft Gaussian Gating} mechanism.}
\vspace{-0.5em}
\label{fig:method}
\end{figure*}

\section{Methodology}
\label{sec:methodology}

In this section, we present \textbf{M}ass-\textbf{A}daptive \textbf{S}oft \textbf{P}olicy \textbf{O}ptimization (MASPO). As illustrated in Figure \ref{fig:intro}, standard RLVR paradigms like GRPO rely on fragmented constraints that create three fundamental misalignments in LLM optimization:

\begin{itemize}[leftmargin=0.5cm]
\item \textbf{Inefficient Gradient Utilization:} Hard clipping imposes a binary cutoff that discards valuable directional gradients from exploratory samples exceeding the boundary, thereby significantly diminishing the effective utilization of informative gradient signals, consequently slowing down the overall convergence speed.
\item \textbf{Insensitive Probability Mass:} Uniform ratio constraints ignore the vast disparity in token probabilities, failing to account for the massive mass displacement in head tokens versus the negligible shift in tail tokens.
\item \textbf{Asymmetric Signal Reliability:} Symmetric advantage handling ignores the disparate signal-to-noise ratios between verified positive solutions and ambiguous negative ones.
\end{itemize}

To address these challenges holistically, MASPO proposes a \textbf{unified} framework (see Figure \ref{fig:method}). We replace the rigid hard clipping with a \textbf{Soft Gaussian Gating} mechanism to ensure gradient continuity (\S\ref{sec:soft_gating}). Crucially, we regulate this gate via a \textbf{Dual-Variable Adaptive Variance} (\S\ref{sec:dual_var}), which synergizes a \textbf{Mass-Adaptive Limiter} for distribution stability and an \textbf{Asymmetric Risk Controller} for signal-aware optimization, ultimately enabling more robust and efficient policy learning.

\subsection{Soft Gaussian Gating}
\label{sec:soft_gating}

To resolve the gradient discontinuity caused by hard clipping, we formulate the MASPO objective using a differentiable confidence score derived from the Principle of Maximum Entropy. The objective is defined as:
\begin{flalign}
\label{equ:maspo_loss}
&\mathcal{J}_{\text{MASPO}}(\theta) = \mathbb{E} _{q \sim \mathcal{D}, \{o_i\}_{i=1}^G \sim \pi_{\theta_{\text{old}}}(\cdot|q)} \frac{1}{\sum_{i=1}^G |o_i|} \nonumber\\ 
&\sum_{i=1}^G \sum_{t=1}^{|o_i|} \mathcal{F}^\text{MASPO}_{i,t} \cdot \rho_{i,t}(\theta) \hat{A}_{i},
\end{flalign}

Here, $\mathcal{F}^\text{MASPO}_{i,t}$ is the \textbf{Soft Gaussian Gating} factor. Unlike hard clipping or SAPO's dual-clip~\cite{ye2020mastering}, this \textbf{unilateral design} selectively attenuates aggressive overshoots without hindering conservative updates of lagging tokens:
\begin{flalign}
\small
\mathcal{F}^\text{MASPO}_{i,t} =
\begin{cases}
\exp \left( {-\frac{(\text{sg}[\rho_{i,t}(\theta)]-1)^{2}}{2 \sigma^{2}_\text{pos}}} \right) & \begin{aligned}[t]
& \text{if } \hat{A}_{i,t}>0 \\
& \quad \land \rho_{i,t}(\theta)>1
\end{aligned} \\
\exp \left( {-\frac{(\text{sg}[\rho_{i,t}(\theta)]-1)^{2}}{2 \sigma^{2}_\text{neg}}} \right) & \begin{aligned}[t]
& \text{if } \hat{A}_{i,t}<0 \\
& \quad \land \rho_{i,t}(\theta)<1
\end{aligned} \\
1, & \text{otherwise}. \\
\end{cases}
\end{flalign}
The stop-gradient operator $\text{sg}[\cdot]$ ensures that $\mathcal{F}$ serves strictly as a confidence gate, effectively preventing the optimization objective from drifting. Mathematically, this transformation converts the optimization landscape from a disjoint cliff (characteristic of binary clipping) into a smooth and continuous manifold. Consequently, samples that marginally exceed the trust region contribute attenuated but non-zero gradients, ensuring that the policy can still learn from ``near-boundary'' explorations without destabilizing the update dynamics.

\subsection{Dual-Variable Adaptive Variance}
\label{sec:dual_var}

The core innovation of MASPO lies in how the variance $\sigma^2$ is dynamically determined. We decouple $\sigma$ into two components (Eq. \eqref{equ:sigma_pos} and \eqref{equ:sigma_neg}) to address the remaining two challenges: insensitive probability mass and asymmetric signal reliability.
\begin{equation}
\label{equ:sigma_pos}
\sigma_\text{pos} = \underbrace{\frac{\sigma_\text{base}}{\pi_{\theta_{old}}^{\alpha}}}_{\text{Mass-Adaptive}} \cdot \underbrace{\left( 1+\beta_\text{high} \hat{A}_{i,t} \right)}_{\text{Risk Controller}},
\end{equation}
\begin{equation}
\label{equ:sigma_neg}
\sigma_\text{neg} = \underbrace{\frac{\sigma_\text{base}}{\pi_{\theta_{old}}^{\alpha}}}_{\text{Mass-Adaptive}} \cdot \underbrace{\left(1-\beta_\text{low} \hat{A}_{i,t} \right)^{-1}}_{\text{Risk Controller}}.
\end{equation}

\paragraph{The Mass-Adaptive Limiter.}
The term $\frac{\sigma_\text{base}}{\pi_{\theta_{old}}^{\alpha}}$ serves as the \textbf{Mass-Adaptive Limiter}. It inversely scales the trust region width with the token's reference probability.
\begin{itemize}[leftmargin=0.5cm]
\item \textbf{In the Long Tail ($\pi \to 0$):} The variance $\sigma$ increases, widening the gate. This compensates for the negligible mass displacement of rare tokens, effectively expanding the exploration budget where it is most needed.
\item \textbf{In the Head ($\pi \to 1$):} The variance shrinks, enforcing strict constraints to prevent policy collapse into a deterministic state due to excessive mass shifts. This aligns with the intuition that for high-confidence tokens, even small ratio deviations imply large absolute mass displacement, necessitating tighter variance to preserve the stability of the generation distribution.
\end{itemize}

\paragraph{The Asymmetric Risk Controller.}
The second component modulates the trust region based on the feedback signal, functioning as an \textbf{Asymmetric Risk Controller}. We derive this design from the monotonicity properties of group-relative advantages (see Appendix \ref{app:monotonicity}), which link advantage magnitude to query difficulty.

\begin{itemize}[leftmargin=0.5cm]
\item \textbf{Positive Signal Expansion ($\sigma_\text{pos}$):} 
For correct reasoning paths ($\hat{A} > 0$), we aim to maximize the utilization of verified signals. Based on the monotonicity property, difficult queries (where correct responses are rare) yield significantly larger positive advantages. These are high-value learning signals. Our expansion term $(1+\beta_\text{high} \hat{A}_{i,t})$ sharpens this effect, allowing the model to take aggressively larger update steps for these high-confidence successes, accelerating the acquisition of complex reasoning skills.

\item \textbf{Negative Signal Conservatism ($\sigma_\text{neg}$):} 
For incorrect paths ($\hat{A} < 0$), the handling is nuanced. According to GRPO monotonicity, ``easy'' queries (where most responses are correct) result in negative advantages with very large magnitudes (e.g., $A^- \ll -1$). In such easy tasks, a negative response is often a ``near-miss'' anomaly where the reasoning chain largely overlaps with correct paths but fails due to a minor error. Applying a massive penalty to these chains risks destroying valid intermediate reasoning patterns crucial for generalization.

MASPO strategically addresses this via the compression term $(1-\beta_\text{low} \hat{A}_{i,t})^{-1}$. When $\hat{A}$ is large and negative, this term significantly shrinks $\sigma_\text{neg}$. This tightly constrains the update for failures in easy tasks, preventing ``catastrophic unlearning'' due to credit assignment ambiguity. Conversely, for hard tasks where failures are common (small negative advantage), the constraint remains relaxed, allowing normal optimization to proceed without excessive restriction.
\end{itemize}
\vspace{-5pt}

\begin{table*}[t]
\small
\centering
\setlength{\tabcolsep}{2.7pt}
\renewcommand{\arraystretch}{1.55}
\caption{Main results on mathematical reasoning benchmarks. We report Avg@32 (A@32) and Pass@32 (P@32). \textbf{Bold} indicates the best performance, and \underline{underlined} represents the second-best.}
\begin{tabular}{lcccccccccccccc}
\toprule
& \multicolumn{2}{c}{\textbf{AIME24}} & \multicolumn{2}{c}{\textbf{AIME25}} & \multicolumn{2}{c}{\textbf{AMC23}} & \multicolumn{2}{c}{\textbf{MATH500}} & \multicolumn{2}{c}{\textbf{Minerva}} & \multicolumn{2}{c}{\textbf{Olympiad}} & \multicolumn{2}{c}{\textbf{Average}} \\
\cmidrule(lr){2-3} \cmidrule(lr){4-5} \cmidrule(lr){6-7} \cmidrule(lr){8-9} \cmidrule(lr){10-11} \cmidrule(lr){12-13} \cmidrule(lr){14-15}
\textbf{Method} & A@32 & P@32 & A@32 & P@32 & A@32 & P@32 & A@32 & P@32 & A@32 & P@32 & A@32 & P@32 & A@32 & P@32 \\

\midrule
\multicolumn{15}{c}{\textit{Backbone: DeepSeek-R1-Distill-Qwen-1.5B}} \\
\midrule

GRPO         & 33.2 & 71.8 & 27.7 & 49.9 & 79.5 & 94.8 & 77.6 & 90.8 & 26.1 & 48.8 & 46.3 & 64.7 & 48.4 & 70.1 \\
Clip Higher  & 36.6 & 70.1 & \best{30.1} & \second{55.8} & \best{82.8} & \second{94.9} & 71.7 & 88.6 & 24.8 & 49.3 & 42.0 & 61.6 & 48.0 & 70.1 \\
DAC          & \second{40.0} & 73.6 & \second{28.7} & 51.2 & 80.2 & \best{95.0} & 78.1 & \best{92.5} & 27.4 & \second{54.0} & 46.3 & 64.5 & 50.1 & \second{71.8} \\
Entropy Adv. & 29.6 & 67.8 & 23.9 & 44.2 & 77.1 & 94.1 & \second{78.5} & 89.7 & 25.2 & 48.6 & 46.4 & 64.3 & 46.8 & 68.1 \\
BAPO         & 38.3 & 72.3 & 28.0 & 55.7 & 80.1 & \second{94.9} & 74.9 & 91.0 & 25.1 & 44.7 & 44.0 & 61.3 & 48.4 & 70.0 \\
SAPO         & 39.3 & \second{73.7} & 28.0 & 49.9 & \second{82.7} & 94.8 & \best{79.1} & \second{91.4} & \second{29.6} & 52.7 & \best{48.2} & \best{66.7} & \second{51.2} & 71.5 \\
\rowcolor{ourshighlight}
\textbf{MASPO} & \best{41.0} & \best{74.8} & 28.4 & \best{58.0} & 82.2 & \best{95.0} & 78.0 & 89.7 & \best{30.7} & \best{54.1} & \second{47.8} & \second{65.7} & \best{51.4} & \best{72.9} \\

\midrule
\multicolumn{15}{c}{\textit{Backbone: DeepSeek-R1-Distill-Qwen-7B}} \\
\midrule

GRPO         & 48.2 & \best{82.5} & 37.4 & 60.5 & 88.1 & 96.6 & 84.8 & 92.4 & 37.4 & 57.2 & 57.2 & 73.9 & 58.9 & 77.2 \\
Clip Higher  & 47.4 & \second{82.4} & 37.4 & \second{67.2} & 89.0 & 96.6 & 85.3 & \best{95.4} & 38.8 & 58.5 & 57.5 & \second{75.5} & \second{59.2} & \second{79.3} \\
DAC          & \second{50.2} & 81.2 & 36.2 & 61.7 & 88.1 & \best{99.6} & 85.1 & \second{94.8} & 36.6 & \best{58.7} & \second{57.9} & \best{76.0} & 59.0 & 78.7 \\
Entropy Adv. & 49.1 & 81.5 & 34.4 & 58.1 & 87.7 & 96.6 & 84.8 & 92.5 & 36.5 & 54.0 & 56.4 & 73.7 & 58.2 & 76.1 \\
BAPO         & 47.1 & 80.3 & \second{37.7} & 58.4 & \second{89.2} & \second{97.2} & 85.0 & 94.5 & 38.3 & 57.5 & 57.1 & 74.5 & 59.1 & 77.1 \\
SAPO         & 47.5 & 79.2 & 35.3 & 58.0 & 88.7 & 95.0 & \second{85.5} & 92.9 & \best{39.4} & 58.1 & 56.6 & 74.6 & 58.8 & 76.3 \\
\rowcolor{ourshighlight}
\textbf{MASPO} & \best{53.2} & \second{82.4} & \best{42.9} & \best{73.2} & \best{91.4} & 95.0 & \best{86.0} & 94.7 & \second{39.3} & \second{58.6} & \best{58.0} & 74.9 & \best{61.8} & \best{79.8} \\

\bottomrule
\end{tabular}
\vspace{-8pt}
\label{tab:main_results}
\end{table*}

\section{Experiments}
\subsection{Experimental Setup}
\paragraph{Implementation Details.}
We leverage the DeepSeek-R1-Distill-Qwen series (specifically 1.5B, 7B, and 14B) as our backbone models to assess MASPO. All models are fine-tuned on the DAPO-Math-17K dataset~\cite{yu2025dapo}. For reward calculation, we adopt a dual-system approach: \texttt{math\_verify}~\cite{Kydlicek_Math-Verify_Math_Verification} is employed for rule-based reward generation during training, whereas \texttt{prime\_math}~\cite{primemath, lightman2023let} serves as the evaluation metric.

\begin{figure*}[t]
\centering
\includegraphics[width=\linewidth]{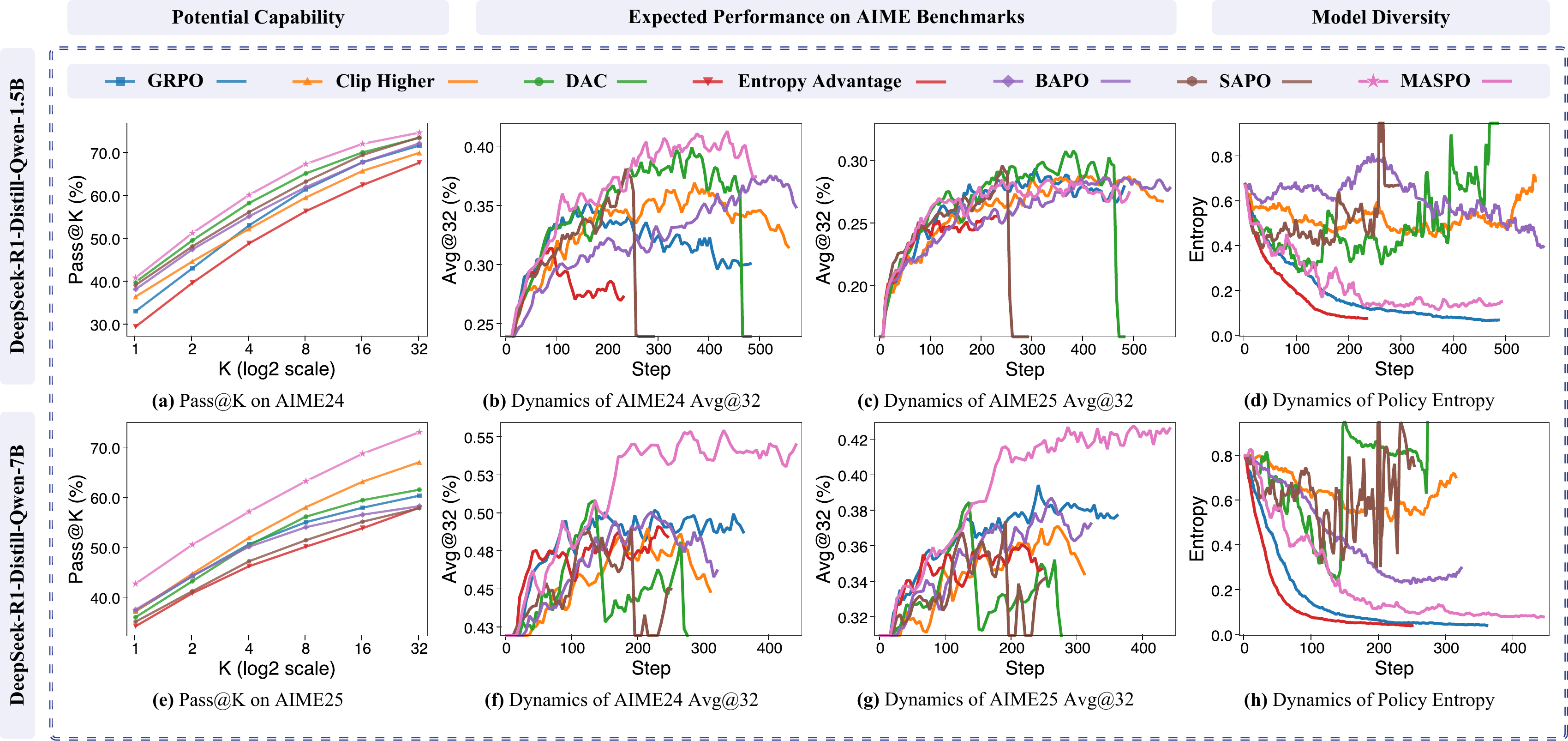}
\caption{Evolution of Training Dynamics and Performance across Model Scales. Top row: 1.5B model; Bottom row: 7B model. MASPO demonstrates superior convergence, achieving higher performance ceilings.}
\label{fig:dynamics}
\vspace{-10pt}
\end{figure*} 

Regarding the optimization dynamics, we configure a global batch size of 512 with a mini-batch size of 32. This setup facilitates 16 off-policy updates per importance sampling step, effectively amplifying the impact of off-policy updates. The clipping thresholds $\varepsilon_\text{low}$ and $\varepsilon_\text{high}$ are both set to 0.2. Additional details are provided in the Appendix \ref{app:configuration}.

\paragraph{Evaluation Protocols.}
To rigorously evaluate the generalization of the model's reasoning capabilities, we conduct evaluations across a diverse set of benchmarks, including AIME24~\cite{AIME24}, AIME25~\cite{AIME25}, AMC23~\cite{AMC}, Minerva~\cite{lewkowycz2022solving}, OlympiadBench~\cite{he2024olympiadbench}, and MATH500~\cite{hendrycks2021measuring}. We report performance using Avg@32 and Pass@32 metrics, which reflect the model's expected stability and its potential peak performance, respectively.

\paragraph{Baseline Methods.}
We benchmark MASPO against a comprehensive suite of strong methods, categorized by their optimization focus: (1) the original \textbf{GRPO} algorithm; (2) \textbf{SAPO}, which removes clipping boundaries entirely and uses a soft weighting function; (3) methods focusing on low-probability compensation, including \textbf{Entropy Regularization} and \textbf{DAC}; (4) approaches emphasizing positive sample rebalancing, such as \textbf{Advantage Reweighting}, \textbf{Entropy Advantage}, and \textbf{BAPO}; and (5) \textbf{Clip Higher}, which implicitly addresses both optimization directions. The comparative results are summarized in Table \ref{tab:main_results}. Note that Entropy Regularization was only preliminarily tested on the 1.5B model, and Advantage Reweighting yielded suboptimal performance; consequently, detailed results for these two methods are omitted from the main table but are provided in Appendix \ref{app:baseline_supp}.

\subsection{Main Results}
\label{sec:main_results}
\paragraph{Overall Performance}
Table \ref{tab:main_results} presents the comparative performance of MASPO against various baselines. MASPO delivers superior performance, achieving well-performing results on the aggregated average across both 1.5B and 7B scales. Specifically, on the 1.5B model, MASPO improves Avg@32 by 3.0\% over GRPO and edges out the closest competitor (SAPO) by 0.2\%. Crucially, this lead expands at the 7B scale. While other baselines plateau near GRPO levels -- with the second-best Clip Higher leading GRPO by only 0.3\% -- MASPO establishes a clear margin, outperforming Clip Higher by 2.6\% in Avg@32. Finally, Pass@K analysis (Figure \ref{fig:dynamics}(a,e)) confirms our method's robustness across difficulty levels: MASPO consistently dominates, leading on AIME24 with the 1.5B model and extending this superiority to the harder AIME25 benchmark at the 7B scale.

\begin{figure*}[t]
    \centering
    \begin{subfigure}[b]{0.32\textwidth}
        \centering
        \includegraphics[width=\linewidth]{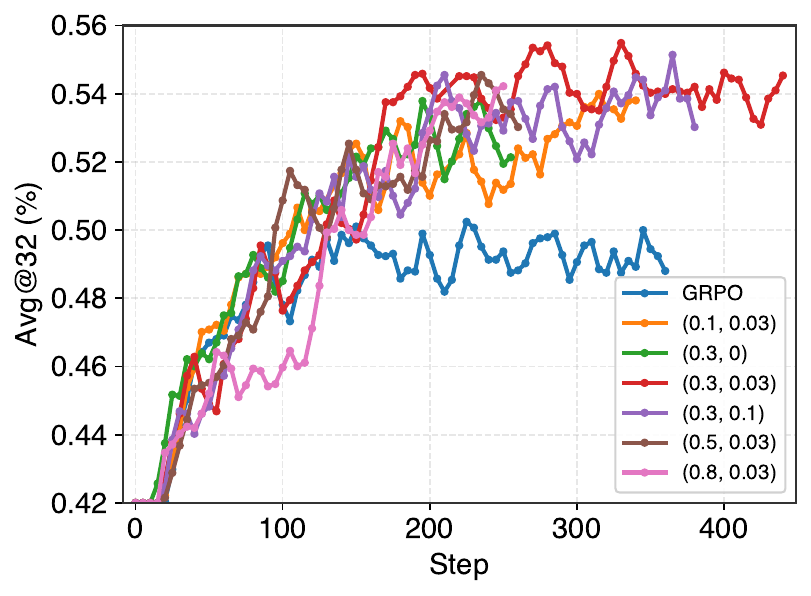}
        \caption{Dynamics of AIME24 Avg@32}
        \label{fig:hyper_aime24}
    \end{subfigure}
    \hfill
    \begin{subfigure}[b]{0.32\textwidth}
        \centering
        \includegraphics[width=\linewidth]{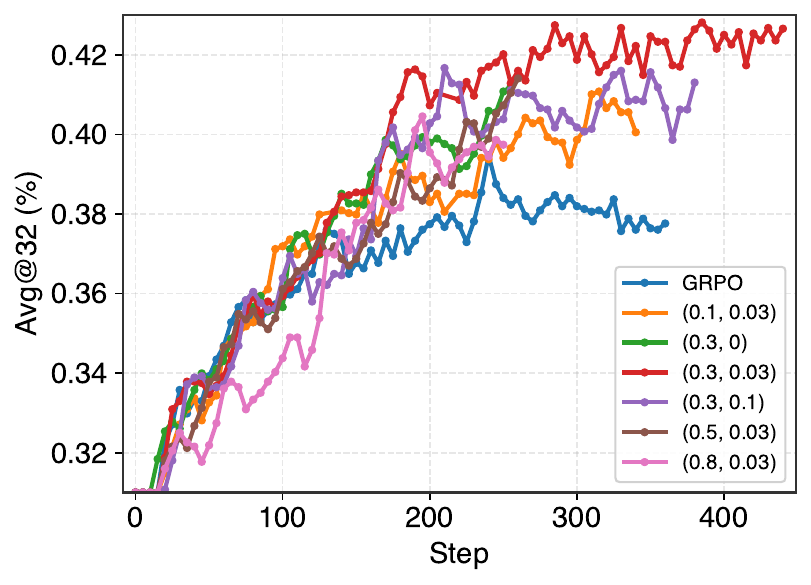}
        \caption{Dynamics of AIME25 Avg@32}
        \label{fig:hyper_aime25}
    \end{subfigure}
    \hfill
    \begin{subfigure}[b]{0.32\textwidth}
        \centering
        \includegraphics[width=\linewidth]{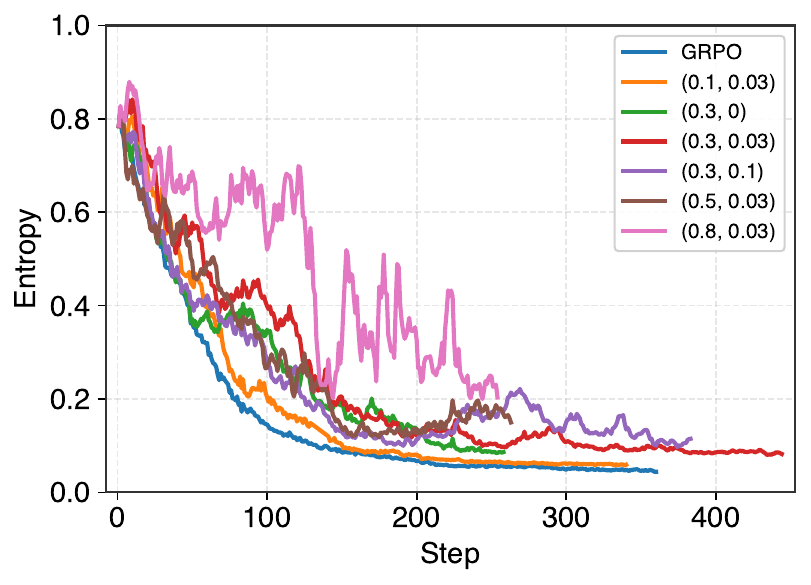}
        \caption{Dynamics of Policy Entropy}
        \label{fig:hyper_entropy}
    \end{subfigure}
    \hfill
    \begin{subfigure}[b]{0.32\textwidth}
        \centering
        \includegraphics[width=\linewidth]{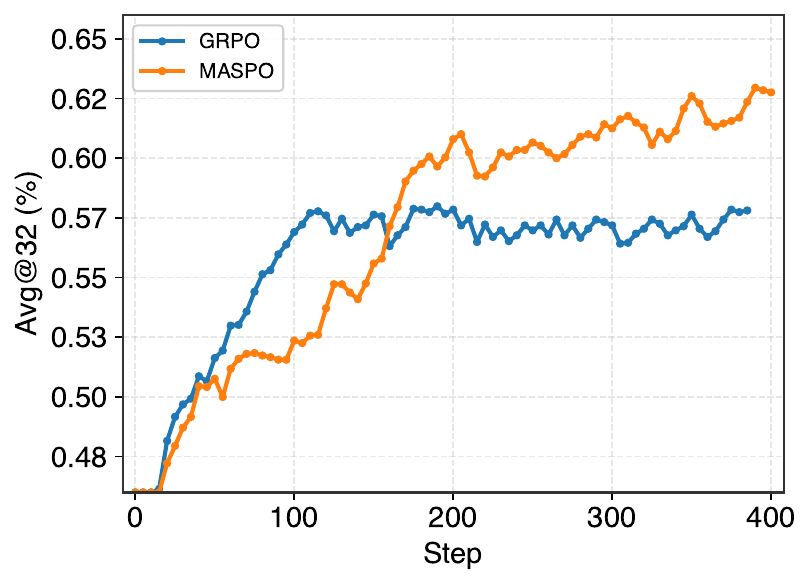}
        \caption{Dynamics of AIME24 Avg@32}
    \end{subfigure}
    \hfill
    \begin{subfigure}[b]{0.32\textwidth}
        \centering
        \includegraphics[width=\linewidth]{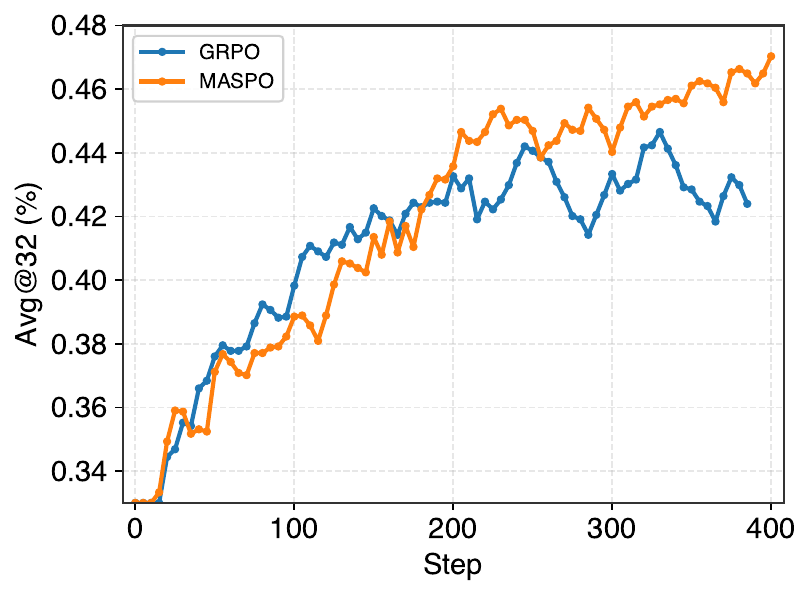}
        \caption{Dynamics of AIME25 Avg@32}
    \end{subfigure}
    \hfill
    \begin{subfigure}[b]{0.32\textwidth}
        \centering
        \includegraphics[width=\linewidth]{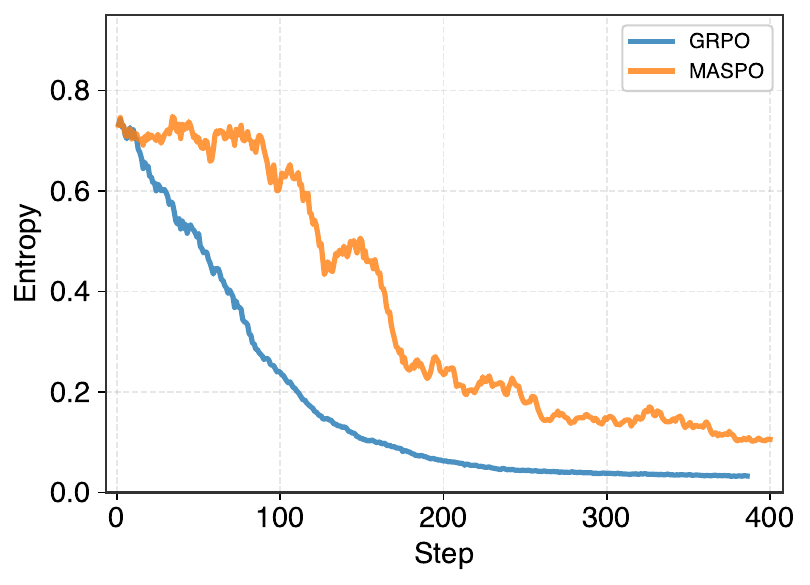}
        \caption{Dynamics of Policy Entropy}
    \end{subfigure}
    \caption{Hyperparameter sensitivity and scalability analysis. \textbf{Top:} MASPO demonstrates robustness across a range of $\alpha$ and $\beta$ values, outperforming the GRPO baseline. \textbf{Bottom:} Training dynamics on the larger 14B model, demonstrating MASPO's scalability by maintaining higher entropy and superior performance compared to GRPO.}
    \label{fig:hyper_dynamics}
    \vspace{-8pt}
\end{figure*}

\paragraph{Training Dynamics.}
The training curves in Figure \ref{fig:dynamics} reveal distinct optimization behaviors. GRPO exhibits a rapid decay in policy entropy, indicating an extreme exploitation mode that sacrifices exploration, consequently limiting its performance ceiling. Entropy Advantage shows an even faster entropy drop, resulting in inferior final performance. 

Among other baselines, Clip Higher, DAC, BAPO, and SAPO maintain higher entropy levels than GRPO, yet this does not strictly translate to better performance. Clip Higher and BAPO suffer from slow policy improvement, only surpassing GRPO on the 1.5B scale. SAPO experiences training collapse on both scales; it only exceeds GRPO's ceiling on the 1.5B model before diverging. We hypothesize this instability stems from its bilateral Gaussian gating design (see the unilateral ablation in Appendix \ref{app:sapo_ablation}). DAC significantly outperforms GRPO on 1.5B but exhibits a ``staircase'' entropy spike on 7B, leading to performance degradation, though it still marginally beats GRPO before regression. In contrast, MASPO maintains a healthy entropy level higher than GRPO without instability or stagnation, achieving a performance ceiling significantly superior to all baselines.

\subsection{Analysis of Hyperparameters}

MASPO introduces four primary hyperparameters: $\sigma_\text{base}$, $\alpha$, $\beta_\text{low}$, and $\beta_\text{high}$. To simplify the analysis, we fix $\sigma_\text{base}=1$ and apply symmetric risk control ($\beta=\beta_\text{low}=\beta_\text{high}$). Using $\alpha=0.3$, $\beta=0.03$ as the baseline configuration, we vary parameters to test the following set: $(0.1, 0.03)$, $(0.3, 0)$, $(0.3, 0.03)$, $(0.3, 0.1)$, $(0.5, 0.03)$, $(0.8, 0.03)$. Experiments were conducted on the 7B model, with results shown in the top row of Figure \ref{fig:hyper_dynamics} and Table \ref{tab:hyper_full}. Our analysis yields the following conclusions (comprehensive data in Appendix \ref{app:hyper_full}):

\paragraph{Robustness.}
As evidenced by the curves in Figure \ref{fig:hyper_dynamics} and Appendix \ref{app:hyper_full}, MASPO demonstrates exceptional robustness. Configurations with $\alpha \in \{0.3, 0.5\}$ yield the optimal performance ceilings. While extreme values ($\alpha=0.1$ or $0.8$) lead to a slight drop, they still consistently outperform GRPO. Furthermore, varying $\beta$ across the tested range maintains superiority over the baseline.

\paragraph{Impact of Mass-Adaptive Scaling ($\alpha$).}
Setting $\alpha=0.1$ imposes excessive constraints, resulting in low entropy levels similar to GRPO and premature loss of exploration capability. Conversely, setting $\alpha=0.8$ overly relaxes constraints, leading to entropy instability and degraded performance. Based on supplementary experiments in Appendix \ref{app:hyper_full}, we recommend $\alpha \in [0.3, 0.5]$ as the optimal range for balancing stability and exploration.

\paragraph{Impact of Asymmetric Risk Control ($\beta$).}
Using the baseline $\alpha=0.3$, setting $\beta=0$ (disabling risk control) causes entropy to drop too rapidly in the early stages, curtailing exploration and slightly lowering the performance ceiling. Increasing $\beta$ to $0.1$ yields a similar effect of premature entropy decay. We recommend $\beta=0.03$ as the optimal setting when $\alpha=0.3$. 

While the efficacy of risk control on negative samples is theoretically justified by the misalignment between advantage and signal-to-noise ratio (see derivation in Appendix \ref{app:monotonicity}), the benefit of sharpening positive advantages requires empirical validation. Our decoupling experiment (Appendix \ref{app:pos_risk_ablation}) confirms that applying risk control to positive samples is crucial for sustaining long-term exploration and preventing performance regression.

\subsection{Scaling Analysis}

To further investigate the scalability of our approach, we applied MASPO to the larger DeepSeek-R1-Distill-Qwen-14B model using $\alpha=0.5$ and $\beta=0.03$. The bottom row of Figure \ref{fig:hyper_dynamics} illustrates the training dynamics. MASPO maintains higher entropy and achieves a significant performance margin over the GRPO baseline. The relative improvements of MASPO across different model scales are summarized in Table \ref{tab:scaling_results} (see Appendix \ref{app:scalability_and_passk} for more details). It is worth noting that the 1.5B and 7B versions are distilled from Qwen-2.5-Math, providing a stronger initial mathematical capability compared to the 14B version, which is based on Qwen-2.5-General. This distinction accounts for the baseline performance variance observed in our scaling experiments.

\section{Conclusion}
In this work, we identify and address structural disconnects in current RLVR paradigms, specifically the inefficiencies from rigid, uniform, and symmetric trust region constraints. We argue prevailing ``hard clipping'' mechanisms fail to accommodate complex dynamics of LLM optimization, particularly regarding the heavy-tailed nature of token distributions and the disparate reliability of reward signals. To bridge these gaps and establish a unified theoretical framework, we propose \textbf{M}ass-\textbf{A}daptive \textbf{S}oft \textbf{P}olicy \textbf{O}ptimization (MASPO). 

\begin{table}[t]
\centering
\small
\renewcommand{\arraystretch}{1.4} 
\setlength{\tabcolsep}{7pt}
\caption{Scalability analysis of average Avg@32 (A) and Pass@32 (P) across 1.5B, 7B, and 14B models.}
\begin{tabular}{lcccccc}
\toprule
\multirow{2}{*}{\textbf{Scale}} & \multicolumn{2}{c}{\textbf{GRPO}} & \multicolumn{2}{c}{\textbf{MASPO}} & \multicolumn{2}{c}{\textbf{Gain \textuparrow}} \\
\cmidrule(lr){2-3} \cmidrule(lr){4-5} \cmidrule(lr){6-7}
 & A & P & A & P & A & P \\
\midrule
1.5B & 48.4 & 70.1 & \best{51.4} & \best{72.9} & \textcolor{teal}{+3.0} & \textcolor{teal}{+2.8}\\
7B   & 58.9 & 77.2 & \best{61.8} & \best{79.8} & \textcolor{teal}{+2.9} & \textcolor{teal}{+2.6}\\
14B  & 53.6 & 67.4 & \best{56.4} & \best{71.1} & \textcolor{teal}{+2.8} & \textcolor{teal}{+3.7}\\
\bottomrule
\end{tabular}
\label{tab:scaling_results}
\vspace{-12pt}
\end{table}

By unifying a differentiable Soft Gaussian Gating mechanism with a Mass-Adaptive Limiter and an Asymmetric Risk Controller, MASPO creates a continuous and responsive optimization landscape. This framework ensures that exploration budgets are dynamically allocated based on token probability mass and that update magnitudes are modulated by the signal-to-noise ratio of the verification feedback. Extensive evaluations across the DeepSeek-R1-Distill-Qwen series (1.5B, 7B, and 14B) confirm that MASPO significantly enhances sample efficiency and reasoning accuracy. Our findings suggest that moving beyond rigid box constraints toward adaptive, probability-aware, and risk-sensitive optimization is a critical step for the next generation of reasoning models.

\section*{Limitations}
We must acknowledge two primary limitations in our current study. (1) \textbf{Dependence on Verifiable Rewards:} Our Asymmetric Risk Controller relies on the premise that positive rewards represent verified truths while negative rewards contain ambiguity (credit assignment noise). Consequently, our experiments are concentrated on mathematical reasoning tasks (e.g., AIME, MATH500) where ground truth is deterministic. The applicability of MASPO to domains with subjective or partial rewards (such as creative writing, general chat, or multimodal retrieval)~\cite{HABIT, OFFSET, INTENT} requires further adaptation of the risk control mechanism. (2) \textbf{Computational Scope:} While we demonstrate scalability from 1.5B to 14B parameters, we have not yet verified MASPO on extremely large-scale foundation models (e.g., 70B+ or MoE architectures) due to computational resource constraints. Although the theoretical derivation of mass-adaptive scaling suggests robustness, empirical validation on such scales remains an important avenue for our future research work. 

\section*{Ethical Considerations}
We have carefully considered potential societal and the ethical implications of our research:
\begin{itemize}[leftmargin=0.5cm]
\item \textbf{Research Integrity:} Throughout this study, we have adhered to established ethical guidelines. We ensure that our methodology is reported transparently, with full disclosure of theoretical derivations and implementation details.
\item \textbf{Data Usage and Privacy:} The datasets utilized in our experiments (e.g., DAPO-Math-17k, AIME) are derived from publicly available, peer-reviewed scientific sources. No private, sensitive, or personally identifiable information (PII) was processed or generated during the course of this research. This commitment to responsible data practices is consistent with broader research efforts~\cite{shao2025language, shao2025narrative}.
\item \textbf{Reproducibility:} To foster scientific progress and transparency within the community, we provide comprehensive documentation of our hyperparameter configurations, including the specific tuning ranges for $\alpha$ and $\beta$. 
\item \textbf{Open Access:} In the interest of community collaboration, we have made our code available anonymously for review and commit to fully open-sourcing the complete implementation upon the final acceptance of this paper.
\end{itemize}

\vspace{-2em}
\bibliography{references}

\appendix

\section{Theoretical Motivation and Derivation}

In this appendix, we provide the rigorous theoretical justification for the \textbf{M}ass-\textbf{A}daptive \textbf{S}oft \textbf{P}olicy \textbf{O}ptimization (MASPO) objective. We deliberately deviate from the standard KL-divergence-based derivation used in TRPO and PPO, arguing that a framework combining \textit{Probability Mass Displacement} with \textit{Probabilistic Relaxation} offers a more robust and mathematically consistent foundation for fine-tuning Large Language Models.
 
\subsection{Step 1: Metric Selection}
\label{app:prob}

Standard policy optimization methods constrain the update using the Kullback-Leibler (KL) divergence, $D_{KL}(\pi_{\theta_{\text{old}}} || \pi_\theta)$. While theoretically sound for distribution matching, KL divergence measures the relative information gain, which can be insensitive to the absolute probability mass of tokens.

In the context of LLM fine-tuning, the stability of the generation process is better preserved by bounding the absolute shift in probability assignment. Ideally, this corresponds to the Total Variation (TV) distance:
\begin{equation}
D_{TV}(\pi_\theta, \pi_{\theta_{\text{old}}}) = \frac{1}{2} \sum_{x} | \pi_\theta(x) - \pi_{\theta_{\text{old}}}(x) |.
\end{equation}
However, optimizing the $L_1$ norm directly is computationally challenging due to its non-differentiability at zero. Consequently, we adopt the \textbf{Squared Mass Displacement} (an $L_2$ proxy) as our constraint metric. For a specific token transition, we require the policy shift to satisfy:
\begin{equation}
\label{eq:l2_constraint}
\frac{1}{2} (\pi_\theta - \pi_{\theta_{\text{old}}})^2 \le \delta^2,
\end{equation}
where $\delta$ is a localized trust region budget. By substituting the probability ratio $\rho(\theta) = \pi_\theta / \pi_{\theta_{\text{old}}}$, we can rewrite this constraint in terms of the ratio:
\begin{align}
\frac{1}{2} \pi_{\theta_{\text{old}}}^2 (\rho(\theta) - 1)^2 &\le \delta^2 \nonumber \\
\implies (\rho(\theta) - 1)^2 &\le \frac{2\delta^2}{\pi_{\theta_{\text{old}}}^2}.
\end{align}
\textbf{Theoretical Implication:} This derivation  reveals a fundamental and critical insight: to maintain a constant safety margin in terms of probability mass ($\delta$), the permissible deviation of the ratio $(\rho-1)$ must be \textit{inversely proportional} to the reference probability $\pi_{\theta_{\text{old}}}$. This finding theoretically justifies the base scaling term in MASPO:
\begin{equation}
\sigma_{\text{base}} \propto \frac{1}{\pi_{\theta_{\text{old}}}^\alpha}.
\end{equation}
While the strict derivation suggests $\alpha=1$, we introduce $\alpha$ as a hyperparameter to empirically control the sensitivity to the long-tail distribution, acknowledging that the $L_2$ proxy may over-encourage rare tokens in high-dimensional vocabularies.

\subsection{Step 2: MaxEnt-based Soft Relaxation}
\label{app:gaussian}

A core technical challenge in GRPO is the use of ``hard clipping'', which effectively imposes a rigid box constraint on the ratio. This creates a disjoint optimization landscape with zero gradients vanishing outside the clip range. In MASPO, we reformulate the constrained optimization problem as a \textbf{Confidence-Weighted Estimation} problem.

Let the ``cost'' of a policy update for a specific sample be defined by mass displacement violation:
\begin{equation}
\mathcal{C}(\rho) = \frac{(\rho-1)^2}{2\sigma^2},
\end{equation}
where $\sigma$ is the adaptive variance previously derived in Step 1. Instead of enforcing a rigid hard constraint $\mathcal{C}(\rho) \le C$, we interpret this cost as a direct measure of \textbf{Gradient Untrustworthiness}. We seek a weighting function $w(\rho) \in [0, 1]$ that modulates the gradient contribution based on this cost.

According to the \textbf{Principle of Maximum Entropy} ~\cite{jaynes1957information1, jaynes1957information2, zhai2026maximizing}, if we possess information about the expected cost (first moment) but are otherwise ignorant about the distribution, the entropy-maximizing distribution is the exponential family. Specifically, we formulate the problem as finding a weight distribution $w(\rho)$ that maximizes entropy $H(w)$ subject to an expected cost constraint:
\begin{equation}
\begin{aligned}
& \max_{w} \quad H(w) = - \int w(\rho) \log w(\rho) d\rho \\
& \text{s.t.} \quad \mathbb{E}_{w}[\mathcal{C}(\rho)] = \mu.
\end{aligned}
\end{equation}
The solution to this Lagrangian optimization is the Gaussian (or Radial Basis) kernel:
\begin{equation}
w^*(\rho) \propto \exp\left( - \beta \cdot \mathcal{C}(\rho) \right) = \exp\left( - \frac{(\rho-1)^2}{2\sigma^2} \right).
\end{equation}
Here, the Lagrange multiplier $\beta$ is absorbed into $\sigma$.

\textbf{From Distribution to Gradient Weighting:}
We apply this optimal distribution $w^*(\rho)$ as a soft gating mechanism for the policy gradient. Unlike the standard objective which assumes uniform reliability across all samples, MASPO optimizes a \textit{reliability-weighted} lower bound:
\begin{align}
\nabla \mathcal{J}_{\text{MASPO}} \approx \mathbb{E}_{o \sim \pi} \bigg[ & \underbrace{\exp\left( - \frac{(\rho-1)^2}{2\sigma^2} \right)}_{\text{Soft Trust Region}} \nonumber \\
& \cdot \rho(\theta) \hat{A} \cdot \nabla \log \pi_\theta \bigg].
\end{align}
This formulation smoothly down-weights samples that exhibit excessive mass displacement (high variance/low confidence) rather than abruptly discarding them, thereby providing a more differentiable and continuous optimization manifold.

\subsection{Step 3: Asymmetric Risk Modeling}
\label{app:adv}

The final component of the derivation addresses the dynamic nature of the budget $\sigma$. While Step 1 established the dependency on probability mass $\pi_{\theta_{\text{old}}}$, we must also account for the nature of the reward signal in reasoning tasks.

Unlike the standard assumption in TRPO/PPO where advantage estimates are treated symmetrically, we proceed from the empirical observation that the Signal-to-Noise Ratio (SNR) in mathematical reasoning and code generation is highly asymmetric. A positive reward verifies a correct reasoning path, while a negative reward is ambiguous (credit assignment problem). We model this asymmetry by modulating the ``trust budget'' $\sigma$ based on the sign of the advantage $\hat{A}$.

\paragraph{Case 1: Positive Advantage (High SNR).}
When $\hat{A} > 0$, the sample indicates a verifiable success. Empirically, we want to exploit these high-quality signals aggressively. We model this as a linear expansion of the trust region:
\begin{equation}
\sigma_{\text{pos}}(\hat{A}) \propto \frac{1}{\pi_{\theta_{\text{old}}}^\alpha} \cdot (1 + \beta_{\text{high}} \hat{A}).
\end{equation}
This formulation allows the policy to take larger steps when the signal is strong and positive, akin to an ``Optimism in the Face of Uncertainty'' strategy.

\paragraph{Case 2: Negative Advantage (Low SNR).}
When $\hat{A} < 0$, the signal is often noisy. An overly aggressive penalty based on a potentially correct intermediate step (labeled negative due to a later error) can lead to unrecoverable policy collapse. Therefore, we treat the negative regime much more conservatively. We use an inverse scaling to tighten the boundary as the negative advantage grows:
\begin{equation}
\sigma_{\text{neg}}(\hat{A}) \propto \frac{1}{\pi_{\theta_{\text{old}}}^\alpha} \cdot \frac{1}{1 - \beta_{\text{low}} \hat{A}}.
\end{equation}
This acts as a safeguard, restricting the magnitude of policy shifts driven by noisy negative feedback.

\textbf{Theoretical Alignment:}
While derived from the specific characteristics of CoT data, this asymmetric variance design aligns with \textbf{Risk-Sensitive Control} theory. It effectively assigns a higher risk cost to updates driven by negative samples (tightening the constraint) while lowering the cost for positive samples (relaxing the constraint), thereby optimizing the exploration-exploitation trade-off in sparse-reward environments.

In the implementation, we use clipping for numerical stability:
\begin{align}
    \sigma_\text{pos} &= \min \left(\frac{\sigma_\text{base}}{\pi_{\theta_{old}}^{\alpha}}, 10\right) \nonumber \\
    &\quad \cdot \ \text{clip} \left(1+\beta_\text{high} \hat{A}_{i,t}, 0.1, 10\right), \\
    \sigma_\text{neg} &= \min \left(\frac{\sigma_\text{base}}{\pi_{\theta_{old}}^{\alpha}}, 10\right) \nonumber \\
    &\quad \cdot \ \text{clip} \left(\frac{1}{1-\beta_\text{low} \hat{A}_{i,t}}, 0.1, 10\right).
\end{align}

\section{Monotonicity Analysis of Group Relative Advantages}
\label{app:monotonicity}

In this section, we analyze the relationship between the advantage values and the difficulty of the task, defined by the number of correct samples within a group. This analysis theoretically supports the asymmetric variance design in MASPO.

Consider a group of size $n$, with $x$ correct samples (positive rewards $r=1$) and $n-x$ incorrect samples (negative rewards $r=-1$). 

First, we calculate the mean reward $\mu$:
\begin{equation}
\mu = \frac{1}{n} \left[ x(1) + (n-x)(-1) \right] = \frac{2x - n}{n}.
\end{equation}

Next, we calculate the variance $\sigma^2$. Note that since $r \in \{-1, 1\}$, $r^2$ is always $1$. Thus, the second moment $\mathbb{E}[r^2] = 1$. The variance is:
\begin{align}
\sigma^2 &= \mathbb{E}[r^2] - (\mathbb{E}[r])^2 = 1 - \left( \frac{2x - n}{n} \right)^2 \nonumber \\
&= 1 - (2p - 1)^2 = 1 - (4p^2 - 4p + 1) \nonumber \\
&= 4p(1-p).
\end{align}
where $p = x/n$ is the accuracy rate. The standard deviation is $\sigma = 2\sqrt{p(1-p)} = \frac{2\sqrt{x(n-x)}}{n}$.

The advantages for positive ($A^+$) and negative ($A^-$) samples are standardized as follows.

For positive samples ($r=1$):
\begin{align}
A^+(x) &= \frac{1 - \mu}{\sigma} = \frac{1 - (2p - 1)}{2\sqrt{p(1-p)}} \nonumber \\
&= \frac{2(1-p)}{2\sqrt{p(1-p)}} = \sqrt{\frac{1-p}{p}} \nonumber \\
&= \sqrt{\frac{n-x}{x}}.
\end{align}

For negative samples ($r=-1$):
\begin{align}
A^-(x) &= \frac{-1 - \mu}{\sigma} = \frac{-1 - (2p - 1)}{2\sqrt{p(1-p)}} \nonumber \\
&= \frac{-2p}{2\sqrt{p(1-p)}} = -\sqrt{\frac{p}{1-p}} \nonumber \\
&= -\sqrt{\frac{x}{n-x}}.
\end{align}

Despite the shift in reward scale from $\{0, 1\}$ to $\{-1, 1\}$, the final expressions for $A^+$ and $A^-$ remain invariant. We now analyze their monotonicity with respect to $x$ (for $x \in (0, n)$):

\paragraph{Positive Advantage ($A^+$).}
\begin{align}
\frac{d A^+}{dx} &= \frac{d}{dx} \left( \frac{n}{x} - 1 \right)^{1/2} \nonumber \\
&= -\frac{n}{2x^2} \sqrt{\frac{x}{n-x}} < 0.
\end{align}
$A^+(x)$ is strictly decreasing with respect to $x$. This implies that for \textbf{hard tasks} (small $x$), the positive advantage is large. MASPO's expansion term $(1+\beta_\text{high} A^+)$ amplifies this, encouraging larger updates for these rare and valuable successes.

\paragraph{Negative Advantage ($A^-$).}
\begin{align}
\frac{d A^-}{dx} &= -\frac{d}{dx} \left( \frac{x}{n-x} \right)^{1/2} \nonumber \\
&= -\frac{n}{2(n-x)^2} \sqrt{\frac{n-x}{x}} < 0.
\end{align}
$A^-(x)$ is strictly decreasing (becoming more negative) as $x$ increases. This implies that for \textbf{easy tasks} (large $x$), the negative advantage has a large absolute magnitude (e.g., $x \to n \implies A^- \to -\infty$). These samples are likely ``near-misses'' in an otherwise easy context. MASPO's compression term $(1-\beta_\text{low} A^-)^{-1}$ becomes very small when $A^-$ is large and negative, effectively suppressing the gradient to prevent excessive penalization of these noisy and low-credit negative samples.

\section{Implementation Details}
\label{app:implementation_details}

\begin{table*}[ht!]
    
    \centering
    \small
    \renewcommand{\arraystretch}{1.7}
    \caption{Summary of the evaluation benchmarks used to assess model performance.}
    \label{tab:benchmark_details}

    \begin{tabularx}{\linewidth}{@{} l X @{}} 
    \toprule
    \textbf{Benchmark} & \textbf{Evaluation Focus \& Characteristics} \\
    \midrule
    \textbf{AIME 2024} & Focuses on arithmetic precision and robustness against contamination. Requires integer outputs (000--999). \\
    \textbf{AIME 2025} & A strictly held-out set representing the frontier of current reasoning capabilities. Tests generalization to unseen, complex logical chains. \\
    \textbf{AMC 2023} & Covers foundational algebra, geometry, and counting. Serves as a baseline for competitive mathematical proficiency. \\
    \textbf{MATH-500} & A representative subset of the MATH dataset, spanning 7 domains including Calculus. Heavily tests LaTeX parsing and symbolic manipulation. \\
    \textbf{Minerva} & Derived from scientific literature, testing higher-order reasoning and domain-specific notation understanding. \\
    \textbf{OlympiadBench} & An aggregate of international competitions (IMO, CMO), representing the upper bound of cross-lingual mathematical logic. \\
    \bottomrule
    \end{tabularx}
    
    \vspace{12pt}

    \centering
    \small
    \caption{Detailed hyperparameter configurations for MASPO and all baselines. The settings listed here correspond to the results reported in the main experimental table.}
    \renewcommand{\arraystretch}{1.4}
    \resizebox{\textwidth}{!}{%
    \begin{tabular}{l l c l}
    \toprule
    \textbf{Category} & \textbf{Algorithm} & \textbf{Learning Rate} & \textbf{Key Hyperparameters} \\
    \midrule
    \multicolumn{4}{l}{\textit{Common Settings: Global Batch Size = 512, Mini-batch Size = 32, Max Length = 8192, $\beta_\text{KL}=0$}} \\
    \midrule
    \multirow{9}{*}{\textbf{DeepSeek-R1-Distill-Qwen-1.5B}} 
     & GRPO & \multirow{9}{*}{$1.0 \times 10^{-6}$} & $\varepsilon=0.2$ \\
     & Entropy Reg. & & $\beta_\text{E}=0.01$ \\
     & Clip Higher & & $\varepsilon_\text{low}=0.2, \varepsilon_\text{high}=0.265$ \\
     & Adv. Reweighting & & $\alpha_\text{A}=0.1$ \\
     & Entropy Adv. & & $\alpha_\text{E}=0.4, \kappa=2.0$ \\
     & DAC & & $\varepsilon_\text{low}=0.2, \varepsilon_\text{high}=0.2$ \\
     & BAPO & & $\begin{aligned} & \rho_0=0.4, a^+=1.2, b^+=3.0, a^-=0.6, \\
     & b^-=0.9, \delta_1=0.05, \delta_2=0.02 \end{aligned}$\\
     & SAPO & & $\tau_\text{neg}=1.05$, $\tau_\text{pos}=1.0$ \\
     & \textbf{MASPO} & & $\sigma_\text{base}=1, \alpha=0.5, \beta=0.03$ \\
    \midrule
    \multirow{7}{*}{\textbf{DeepSeek-R1-Distill-Qwen-7B}} 
     & GRPO & \multirow{7}{*}{$4.63 \times 10^{-7}$} & $\varepsilon=0.2$ \\
     & Clip Higher & & $\varepsilon_\text{low}=0.2, \varepsilon_\text{high}=0.265$ \\
     & Adv. Reweighting & & $\alpha_\text{A}=0.1$ \\
     & Entropy Adv. & & $\alpha_\text{E}=0.4, \kappa=2.0$ \\
     & DAC & & $\varepsilon_\text{low}=0.2, \varepsilon_\text{high}=0.2$ \\
     & BAPO & & $\begin{aligned} & \rho_0=0.4, a^+=1.2, b^+=3.0, a^-=0.6, \\
     & b^-=0.9, \delta_1=0.05, \delta_2=0.02 \end{aligned}$\\
     & SAPO & & $\tau_\text{neg}=1.05$, $\tau_\text{pos}=1.0$ \\
     & \textbf{MASPO} & & $\sigma_\text{base}=1, \alpha=0.3, \beta=0.03$ \\
    \midrule
    \multirow{2}{*}{\textbf{DeepSeek-R1-Distill-Qwen-14B}} 
     & GRPO & \multirow{2}{*}{$3.27 \times 10^{-7}$} & $\varepsilon=0.2$ \\
     & \textbf{MASPO} & & $\sigma_\text{base}=1, \alpha=0.5, \beta_\text{low}=0.03$ \\
    \bottomrule
    \end{tabular}%
    }
    \vspace{-1em}
    \label{tab:hyper_settings}
\end{table*}

\subsection{Experimental Ecosystem}

\paragraph{Training Infrastructure.}
All experiments were orchestrated on a distributed high-performance computing cluster. The hardware topology consists of 30 compute nodes, each populated with 8 $\times$ NVIDIA A100 (80GB) GPUs. Inter-node communication is facilitated by a high-bandwidth InfiniBand fabric, while intra-node data transfer utilizes NVLink to minimize gradient synchronization latency. The cumulative compute budget for this study exceeded 20,000 GPU-hours.

\paragraph{Software Pipeline.}
We implemented MASPO within the VeRL framework~\cite{sheng2025hybridflow}, optimized for post-training large-scale reasoning.
\begin{itemize}[leftmargin=0.5cm]
    \item \textbf{Distributed Strategy:} We employed Fully Sharded Data Parallel (FSDP) to shard parameters, gradients, and optimizer states across ranks. This allows full-parameter fine-tuning of the 14B model without necessitating parameter-efficient approximations like LoRA.
    \item \textbf{Inference Optimization:} To handle extensive rollout generation required by group-based RL, we integrated vLLM with PagedAttention. This setup efficiently manages the Key-Value (KV) cache, enabling high-throughput generation of long chain-of-thought sequences (up to 8192 tokens) with minimal memory fragmentation~\cite{dong2025aurora}.
\end{itemize}

\subsection{Task Definition and Evaluation}

\paragraph{Training Data and Tokenization.}
Our training utilizes the DAPO-Math-17k dataset, which consists entirely of publicly available mathematical problems with verifiable ground-truth answers, ensuring data integrity throughout the pipeline~\cite{liu2025stole}. We strictly adhere to the Qwen2.5-Math tokenizer to maintain alignment between supervised fine-tuning (SFT) and the RLVR stage. We enforce a maximum sequence length of 8192 tokens, a critical setting that permits the model to explore complex, multi-step reasoning paths without artificial truncation.

\paragraph{Reward Mechanism.}
We adopt a sparse, outcome-based reward function. For each query $q$, the generated response $o$ is parsed to extract the final answer. We utilize the \texttt{math\_verify} library for robust answer extraction and equivalence checking.
\begin{equation}
    r(q, o) = \begin{cases} 
    1, & \text{if answer matches ground truth} \\
    -1, & \text{otherwise}.
    \end{cases}
\end{equation}
This binary signal $\{-1, 1\}$ presents a challenging exploration landscape~\cite{liu2023retrieval, liu2024unsupervised}, necessitating the mass-adaptive and risk-sensitive mechanisms embedded within MASPO.

\paragraph{Benchmark Suite.}
To provide a holistic assessment of mathematical reasoning, we evaluate on a diverse set of benchmarks ranging from high school competitions to collegiate-level problems. Detailed characteristics are provided in Table \ref{tab:benchmark_details}.

\subsection{Configuration and Reproducibility}
\label{app:configuration}

\paragraph{Optimization Protocol.}
We utilize the AdamW optimizer with $\beta_1=0.9$ and $\beta_2=0.95$. To isolate the algorithmic contribution of MASPO, we employ a \textbf{constant learning rate} schedule across all experiments, avoiding the confounding effects of warm-up or decay schedules. The global batch size is set to 512 for rollouts and 32 for updates. We set the KL penalty coefficient $\beta_\text{KL}=0$, relying implicitly on the trust region constraints.

\begin{figure*}[ht!]

    \centering

    \begin{subfigure}[b]{0.32\textwidth}
        \centering
        \includegraphics[width=\linewidth]{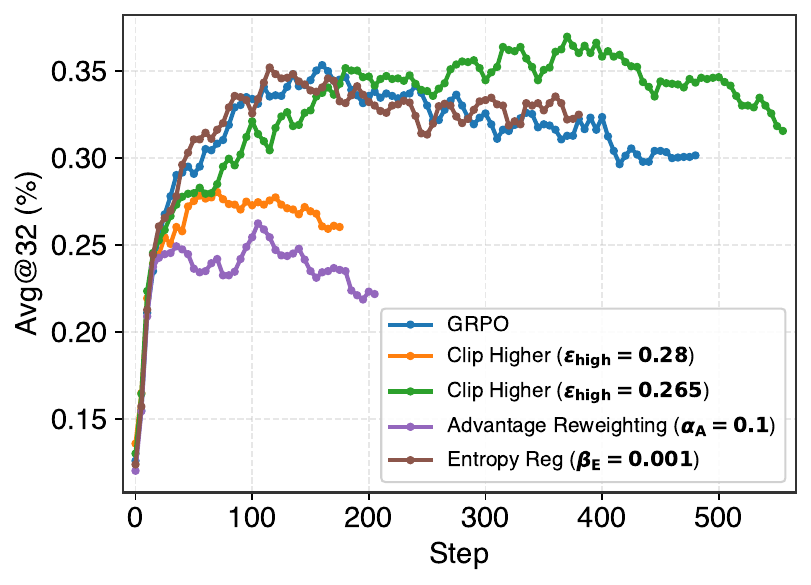}
        \caption{AIME24 Dynamics of 1.5B Model}
    \end{subfigure}
    \hfill
    \begin{subfigure}[b]{0.32\textwidth}
        \centering
        \includegraphics[width=\linewidth]{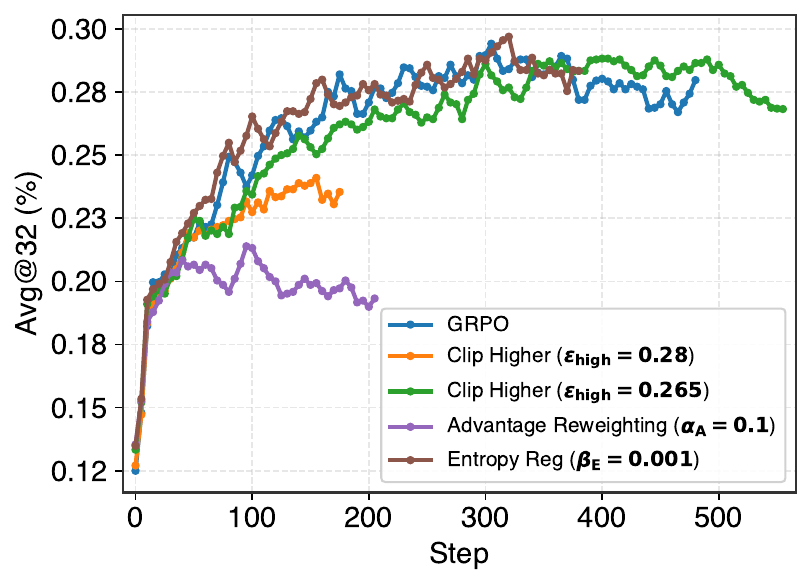}
        \caption{AIME25 Dynamics of 1.5B Model}
    \end{subfigure}
    \hfill
    \begin{subfigure}[b]{0.32\textwidth}
        \centering
        \includegraphics[width=\linewidth]{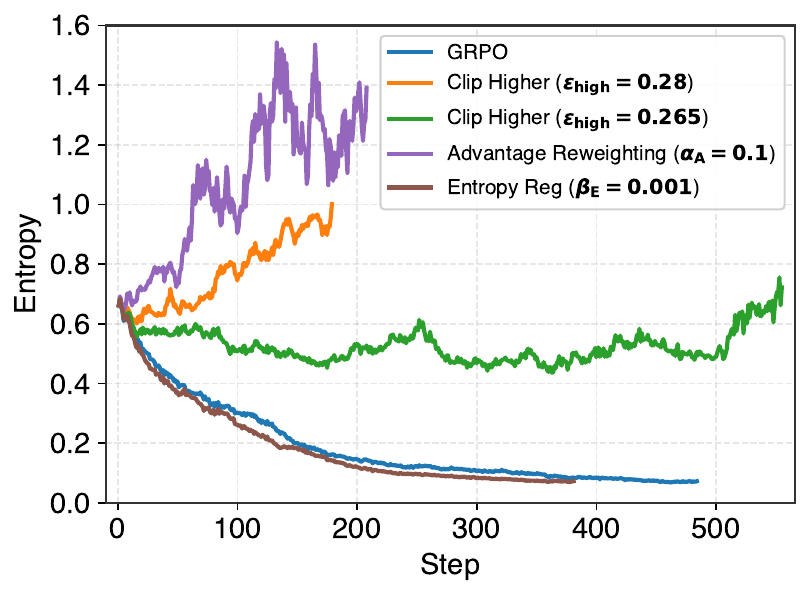}
        \caption{Entropy Dynamics of 1.5B Model}
    \end{subfigure}
    
    \vspace{0.2cm}

    \begin{subfigure}[b]{0.32\textwidth}
        \centering
        \includegraphics[width=\linewidth]{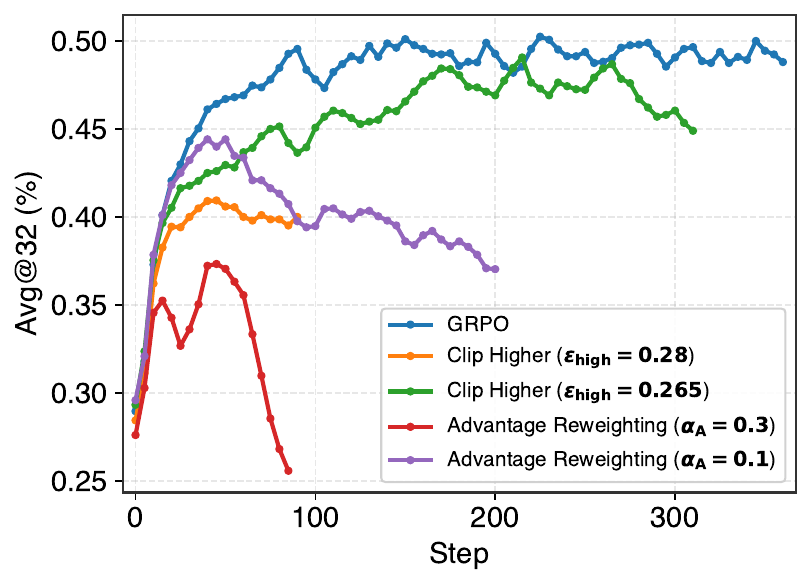}
        \caption{AIME24 Dynamics of 7B Model}
    \end{subfigure}
    \hfill
    \begin{subfigure}[b]{0.32\textwidth}
        \centering
        \includegraphics[width=\linewidth]{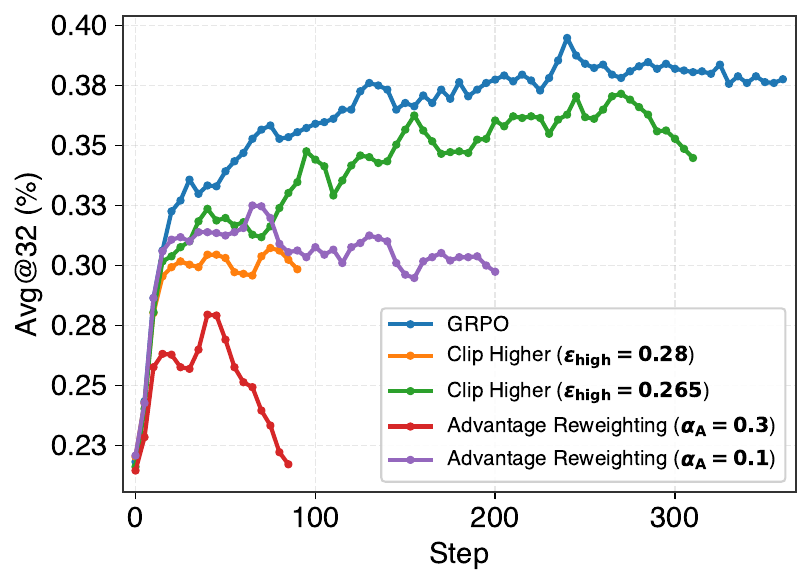}
        \caption{AIME25 Dynamics of 7B Model}
    \end{subfigure}
    \hfill
    \begin{subfigure}[b]{0.32\textwidth}
        \centering
        \includegraphics[width=\linewidth]{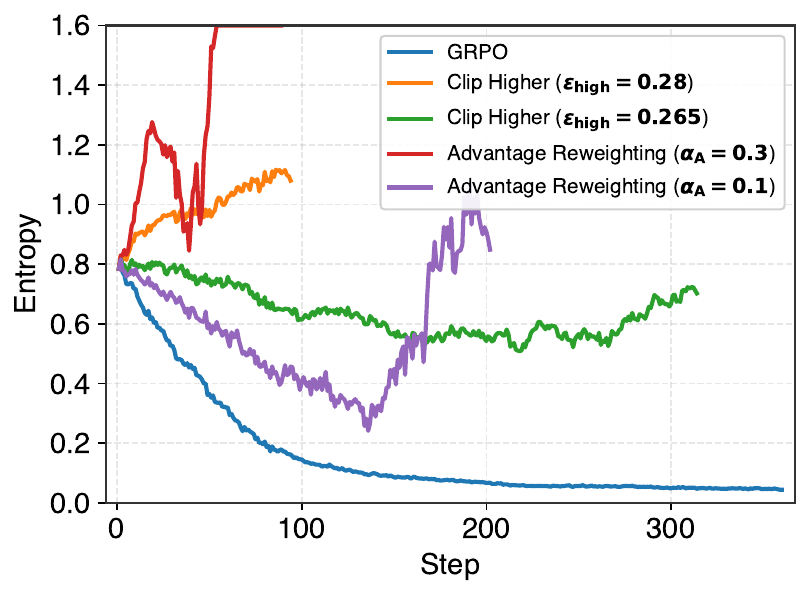}
        \caption{Entropy Dynamics of 7B Model}
    \end{subfigure}
    \caption{Supplementary training dynamics for additional baselines (Clip Higher, Advantage Reweighting, etc.).}
    \label{fig:baseline_supp}
    \vspace{-6pt}

\end{figure*}

\begin{figure*}[b!]
    \centering
    \begin{subfigure}[b]{0.32\textwidth}
        \centering
        \includegraphics[width=\linewidth]{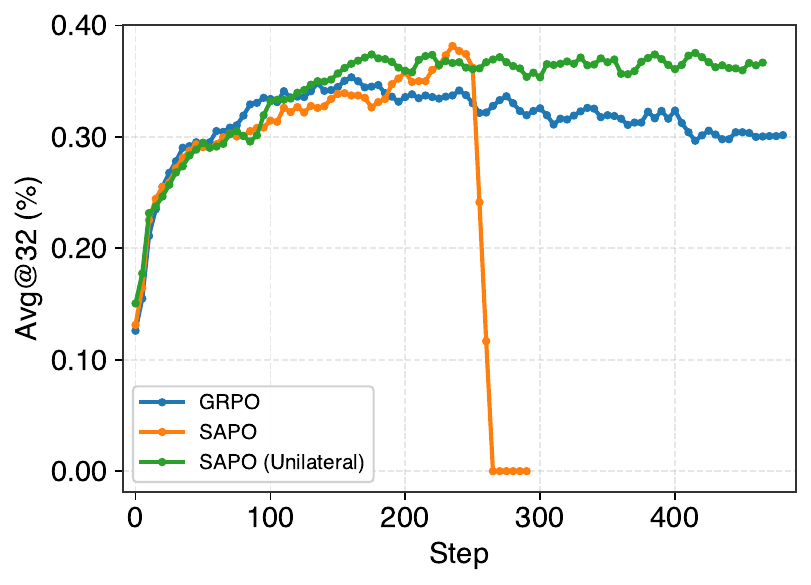}
        \caption{AIME24 Accuracy}
    \end{subfigure}
    \hfill
    \begin{subfigure}[b]{0.32\textwidth}
        \centering
        \includegraphics[width=\linewidth]{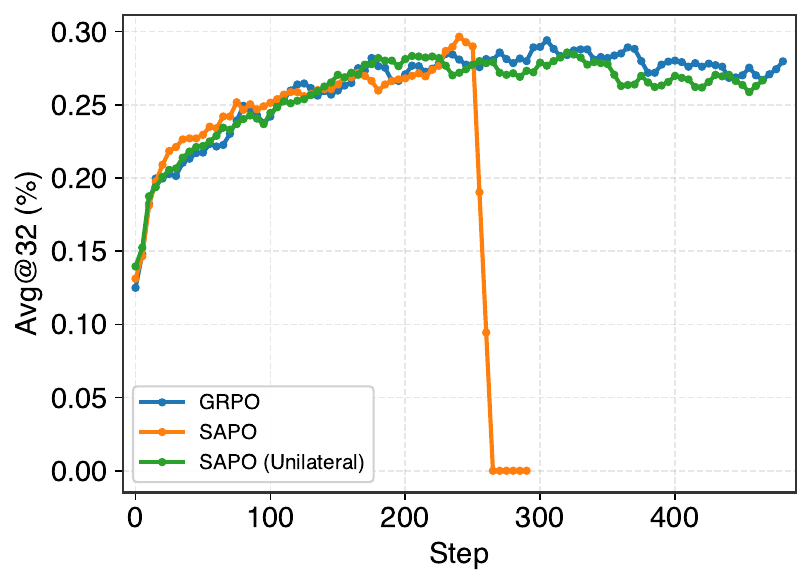}
        \caption{AIME25 Accuracy}
    \end{subfigure}
    \hfill
    \begin{subfigure}[b]{0.32\textwidth}
        \centering
        \includegraphics[width=\linewidth]{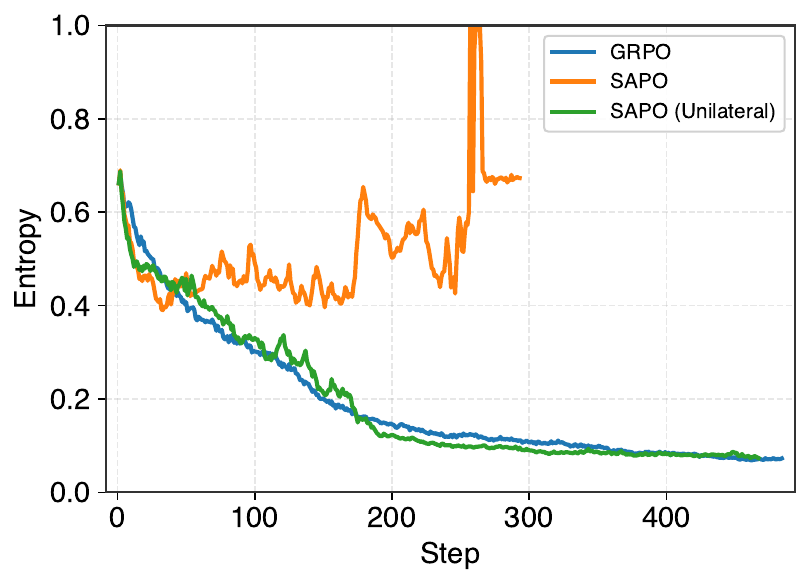}
        \caption{Policy Entropy}
    \end{subfigure}
    \caption{Ablation study of the SAPO gating mechanism on the 1.5B model. The unilateral design (similar to MASPO) prevents the training collapse observed in the original bilateral SAPO.}
    \label{fig:sapo_ablation}
\end{figure*}

\paragraph{Baseline Configurations.}
For all baseline algorithms, we adopt the optimal hyperparameter settings reported in their respective original papers or derived from our best-effort grid search (as detailed in Section \ref{app:baseline_supp}). Table \ref{tab:hyper_settings} lists the specific configurations used for the main results in Table \ref{tab:main_results}.

\section{Supplementary Experimental Analysis}
\label{app:exp_details}

\subsection{Additional Baseline Experiments}
\label{app:baseline_supp}
We provide further context for baseline comparisons. For Clip Higher, the default $\varepsilon_\text{high}=0.28$ performed poorly on both 1.5B and 7B scales (Figure \ref{fig:baseline_supp}). Following prior exploration strategies~\cite{he2025skywork}, we adopted $\varepsilon_\text{high}=0.265$, which yielded the competitive results reported in the main Table \ref{tab:main_results}. For Advantage Reweighting, the default $\alpha_\text{A}=0.3$ underperformed; adjusting to $\alpha_\text{A}=0.1$ improved results, though it still lagged behind GRPO. Experiments with Entropy Regularization on the 1.5B scale showed negligible difference compared to standard GRPO.

\begin{table*}[t]
\vspace{-12pt}
\footnotesize
\centering
\renewcommand{\arraystretch}{1.4} 
\setlength{\tabcolsep}{2pt}
\caption{Full hyperparameters comparison on DeepSeek-R1-Distill-Qwen-7B.}
\begin{tabular}{lcccccccccccccc}
\toprule
\multirow{2}{*}{\textbf{Hyperparameters}} & \multicolumn{2}{c}{\textbf{AIME24}} & \multicolumn{2}{c}{\textbf{AIME25}} & \multicolumn{2}{c}{\textbf{AMC23}} & \multicolumn{2}{c}{\textbf{MATH500}} & \multicolumn{2}{c}{\textbf{Minerva}} & \multicolumn{2}{c}{\textbf{Olympiad}} & \multicolumn{2}{c}{\textbf{Avg.}} \\
\cmidrule(lr){2-3} \cmidrule(lr){4-5} \cmidrule(lr){6-7} \cmidrule(lr){8-9} \cmidrule(lr){10-11} \cmidrule(lr){12-13} \cmidrule(lr){14-15}
& A@32 & P@32 & A@32 & P@32 & A@32 & P@32 & A@32 & P@32 & A@32 & P@32 & A@32 & P@32 & A@32 & P@32 \\
\midrule
$\alpha=0.1$, $\beta=0.03$ & 53.1 & 78.4 & 41.3 & 72.0 & 89.5 & 97.5 & 84.1 & 90.8 & 38.1 & 55.5 & 56.2 & 70.6 & 60.4 & 77.5 \\
$\alpha=0.3$, $\beta=0$    & 54.8 & 79.8 & 39.8 & 58.7 & 90.3 & 96.6 & 84.7 & 92.9 & 38.2 & 55.2 & 57.3 & 73.6 & 60.9 & 76.1 \\
$\alpha=0.3$, $\beta=0.03$ & 53.2 & 82.4 & \textbf{42.9} & \textbf{73.2} & 91.4 & 95.0 & \textbf{86.0} & \textbf{94.7} & 39.3 & 58.6 & 58.0 & \textbf{74.9} & 61.8 & \textbf{79.8} \\
$\alpha=0.3$, $\beta=0.1$  & 55.3 & \textbf{85.6} & 40.1 & 66.3 & 90.9 & \textbf{98.7} & 84.1 & 92.9 & 38.2 & 55.0 & 55.8 & 71.4 & 60.7 & 78.3 \\
$\alpha=0.5$, $\beta=0$    & 51.5 & 81.8 & 41.6 & 64.6 & 90.5 & 97.2 & 84.4 & 91.6 & 38.8 & 55.9 & 56.5 & 71.6 & 60.6 & 77.1 \\
$\alpha=0.5$, $\beta=0.03$ & 55.0 & 82.4 & 42.5 & 65.2 & 91.6 & 98.3 & 85.3 & 92.3 & 39.3 & 55.2 & \textbf{58.1} & 72.3 & \textbf{62.0} & 77.6 \\
$\alpha=0.5$, $\beta=0.1$  & \textbf{56.3} & 85.2 & 39.7 & 60.8 & 91.0 & 95.0 & 83.1 & 93.5 & 37.8 & \textbf{58.8} & 54.8 & 71.8 & 60.5 & 77.5 \\
$\alpha=0.8$, $\beta=0.03$ & 54.9 & 82.7 & 40.5 & 64.8 & \textbf{92.0} & 97.5 & 84.8 & 92.3 & \textbf{40.1} & 56.4 & 56.0 & 72.6 & 61.4 & 77.7 \\
\bottomrule
\end{tabular}
\label{tab:hyper_full}
\vspace{2pt}
\end{table*}

\begin{figure*}[ht!]
    \centering
    \begin{subfigure}[b]{0.32\textwidth}
        \centering
        \includegraphics[width=\linewidth]{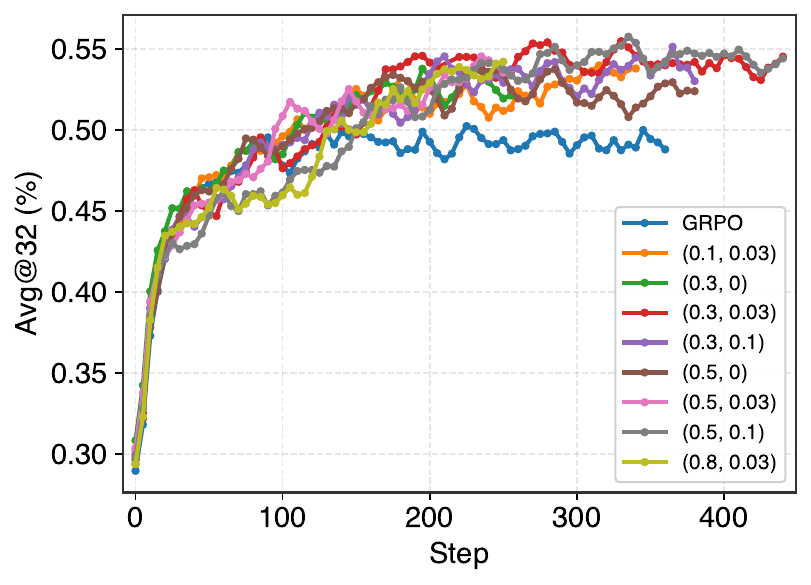}
        \caption{Dynamics of AIME24 Accuracy}
    \end{subfigure}
    \hfill
    \begin{subfigure}[b]{0.32\textwidth}
        \centering
        \includegraphics[width=\linewidth]{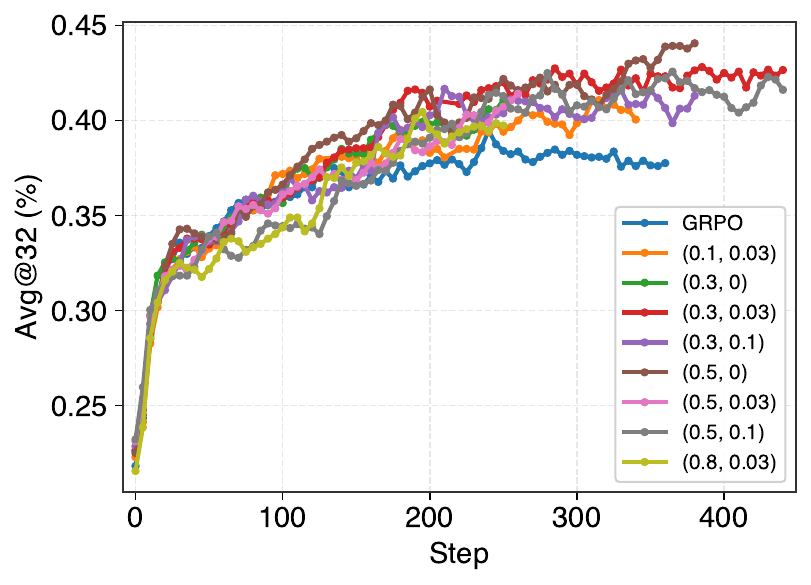}
        \caption{Dynamics of AIME25 Accuracy}
    \end{subfigure}
    \hfill
    \begin{subfigure}[b]{0.32\textwidth}
        \centering
        \includegraphics[width=\linewidth]{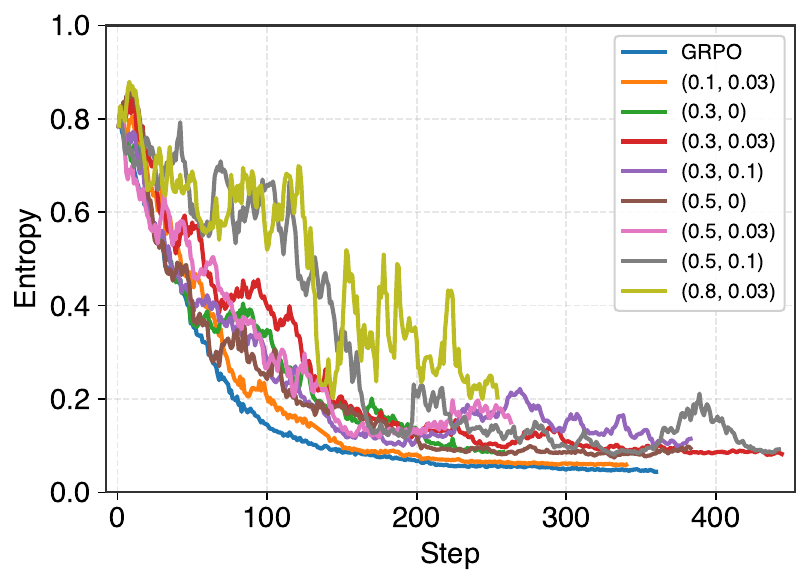}
        \caption{Dynamics of Policy Entropy}
    \end{subfigure}
    \caption{Complete hyperparameter sweep results on the DeepSeek-R1-Distill-Qwen-7B, confirming the robustness of MASPO across a wider range of hyperparameter settings.}
    \label{fig:hyper_full}

    \vspace{10pt}

    \centering
    \begin{subfigure}[b]{0.32\textwidth}
        \centering
        \includegraphics[width=\linewidth]{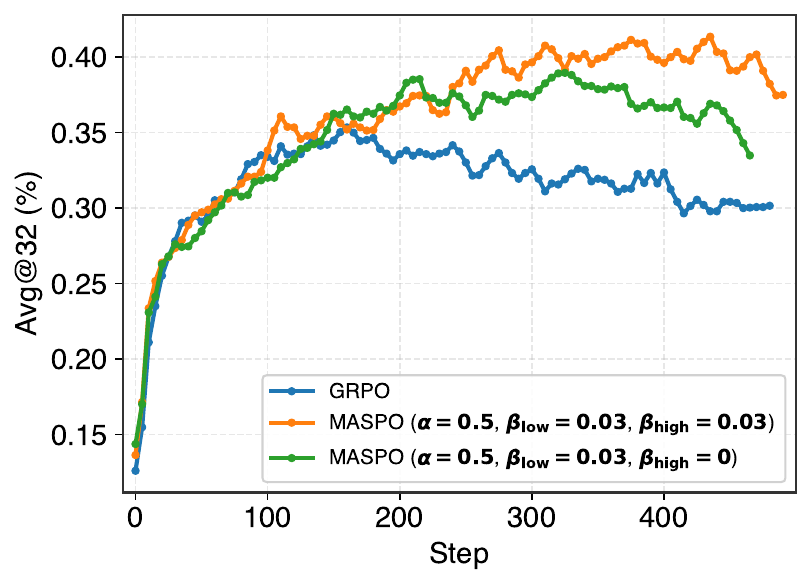}
        \caption{Dynamics of AIME24 Avg@32}
    \end{subfigure}
    \hfill
    \begin{subfigure}[b]{0.32\textwidth}
        \centering
        \includegraphics[width=\linewidth]{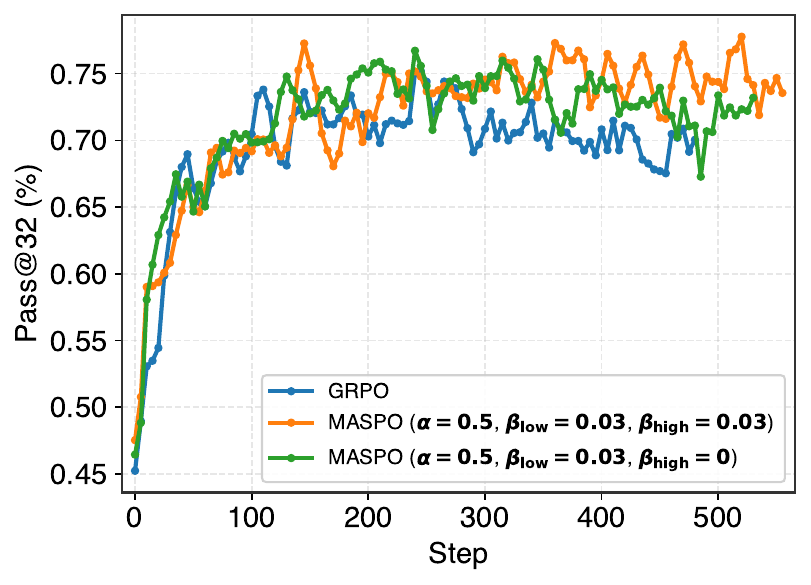}
        \caption{Dynamics of AIME24 Pass@32}
    \end{subfigure}
    \hfill
    \begin{subfigure}[b]{0.32\textwidth}
        \centering
        \includegraphics[width=\linewidth]{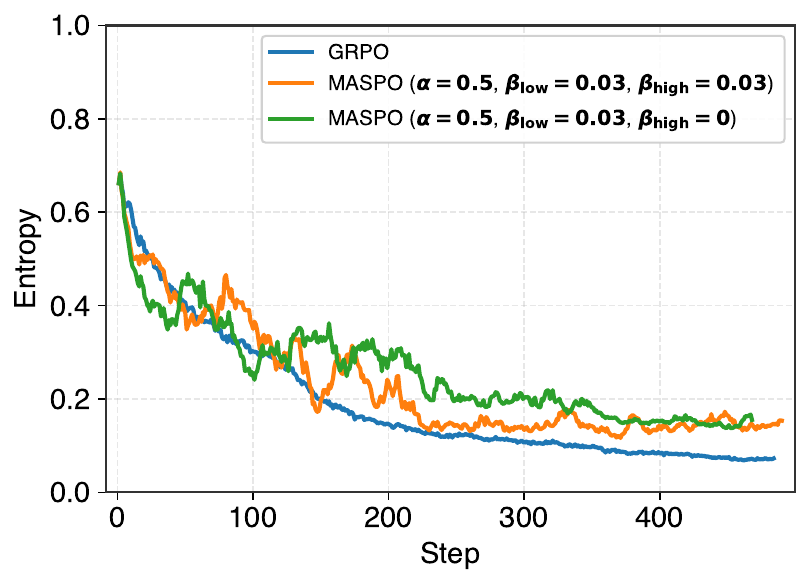}
        \caption{Dynamics of Policy Entropy}
    \end{subfigure}
    \caption{Ablation study on positive risk control ($\beta_\text{high}$) using the 1.5B model.}
    \label{fig:pos_ablation}
    \vspace{-4pt}
\end{figure*}

\subsection{Ablation on SAPO Gating Mechanism}
\label{app:sapo_ablation}

To investigate the instability of SAPO, we modified it to adopt a unilateral gating form identical to MASPO—applying the soft Gaussian gating only when the update is potentially destabilizing, i.e., when $(\hat{A}_{i,t}>0 \land \rho_{i,t}(\theta)>1)$ or $(\hat{A}_{i,t}<0 \land \rho_{i,t}(\theta)<1)$. Unlike SAPO's bilateral design, which indiscriminately attenuates all token updates and suppresses even conservative gradients, the unilateral form preserves beneficial updates. Experiments on DeepSeek-R1-Distill-Qwen-1.5B show that this unilateral version avoids the collapse observed in the original bilateral SAPO. This variant effectively serves as an ablation of MASPO without the Mass-Adaptive Variance and Risk Controller, further validating the rationale behind MASPO's unilateral design.

\subsection{Complete Hyperparameter Results}
\label{app:hyper_full}

Figure \ref{fig:hyper_full} and Table \ref{tab:hyper_full} detail the full hyperparameter sweep on DeepSeek-R1-Distill-Qwen-7B. The results confirm that alongside the primary $\alpha=0.3$, the $\alpha=0.5$ also delivers robust and superiority.

\begin{table*}[htbp]
    \centering
    
    \small
    \setlength{\tabcolsep}{2.8pt}
    \renewcommand{\arraystretch}{1} 
    \caption{Performance results on the DeepSeek-R1-Distill-Qwen-14B model. We report Avg@32 and Pass@32.}
    \label{tab:14b_full_results}
    \begin{tabular}{lcccccccccccccc}
    \toprule
    & \multicolumn{2}{c}{\textbf{AIME24}} & \multicolumn{2}{c}{\textbf{AIME25}} & \multicolumn{2}{c}{\textbf{AMC23}} & \multicolumn{2}{c}{\textbf{MATH500}} & \multicolumn{2}{c}{\textbf{Minerva}} & \multicolumn{2}{c}{\textbf{Olympiad}} & \multicolumn{2}{c}{\textbf{Average}} \\
    \cmidrule(lr){2-3} \cmidrule(lr){4-5} \cmidrule(lr){6-7} \cmidrule(lr){8-9} \cmidrule(lr){10-11} \cmidrule(lr){12-13} \cmidrule(lr){14-15}
    \textbf{Method} & A@32 & P@32 & A@32 & P@32 & A@32 & P@32 & A@32 & P@32 & A@32 & P@32 & A@32 & P@32 & A@32 & P@32 \\
    \midrule
    GRPO & 56.6 & 82.9 & 40.5 & 67.8 & 92.2 & 97.4 & 66.5 & 70.4 & 22.6 & 35.3 & 43.2 & 50.7 & 53.6 & 67.4 \\
    \rowcolor{ourshighlight}
    \textbf{MASPO} & \best{63.5} & \best{86.1} & \best{45.3} & \best{72.4} & \best{93.4} & \best{99.9} & \best{66.8} & \best{73.3} & \best{24.6} & \best{41.4} & \best{45.0} & \best{53.7} & \best{56.4} & \best{71.1} \\
    \bottomrule
    \end{tabular}

    \vspace{15pt}
    
    \small 
    \setlength{\tabcolsep}{15.8pt} 
    \renewcommand{\arraystretch}{1} 
    \caption{Detailed Pass@$k$ performance on AIME 2024 and AIME 2025 across all model scales.}
    \label{tab:pass_k_detailed}
    \begin{tabular}{lcccccc}
    \toprule
    \textbf{Method} & \textbf{Pass@1} & \textbf{Pass@2} & \textbf{Pass@4} & \textbf{Pass@8} & \textbf{Pass@16} & \textbf{Pass@32} \\
    \midrule
    
    \multicolumn{7}{c}{\textit{Backbone: DeepSeek-R1-Distill-Qwen-1.5B (AIME 2024)}} \\
    \midrule
    GRPO         & 33.2 & 43.2 & 53.2 & 61.6 & 67.9 & 71.8 \\
    Clip Higher  & 36.6 & 44.8 & 52.4 & 59.7 & 65.9 & 70.1 \\
    DAC 400      & 40.0 & 49.7 & 58.4 & 65.3 & 70.2 & 73.6 \\
    Entropy Adv  & 29.6 & 39.8 & 49.0 & 56.5 & 62.6 & 67.8 \\
    BAPO         & 38.3 & 47.6 & 55.3 & 62.1 & 67.9 & 72.3 \\
    SAPO         & 39.3 & 48.3 & 56.3 & 63.4 & 69.6 & 73.7 \\
    \rowcolor{ourshighlight}
    \textbf{MASPO} & \best{41.0} & \best{51.4} & \best{60.3} & \best{67.5} & \best{72.2} & \best{74.8} \\
    
    \midrule
    \multicolumn{7}{c}{\textit{Backbone: DeepSeek-R1-Distill-Qwen-1.5B (AIME 2025)}} \\
    \midrule
    GRPO         & 27.7 & 33.0 & 38.1 & 42.6 & 46.5 & 49.9 \\
    Clip Higher  & \best{30.1} & \best{34.8} & \best{39.7} & 45.0 & 50.8 & 55.8 \\
    DAC 400      & 28.7 & 34.1 & 39.0 & 43.6 & 48.0 & 51.2 \\
    Entropy Adv  & 23.9 & 29.0 & 32.9 & 37.0 & 41.1 & 44.2 \\
    BAPO         & 28.0 & 33.0 & 37.5 & 42.8 & 49.3 & 55.7 \\
    SAPO         & 28.0 & 32.1 & 36.0 & 40.8 & 45.7 & 49.9 \\
    \rowcolor{ourshighlight}
    \textbf{MASPO} & 28.4 & 33.9 & 39.3 & \best{45.1} & \best{51.6} & \best{58.0} \\
    
    \midrule
    \multicolumn{7}{c}{\textit{Backbone: DeepSeek-R1-Distill-Qwen-7B (AIME 2024)}} \\
    \midrule
    GRPO         & 48.2 & 58.0 & 66.6 & 73.4 & 78.6 & \best{82.5} \\
    Clip Higher  & 47.4 & 56.7 & 65.0 & 72.6 & 78.4 & 82.4 \\
    DAC 400      & 50.2 & 59.5 & 67.0 & 72.8 & 77.9 & 81.2 \\
    Entropy Adv  & 49.1 & 59.8 & 68.9 & 75.4 & 79.4 & 81.5 \\
    BAPO         & 47.1 & 56.5 & 64.9 & 71.7 & 76.8 & 80.3 \\
    SAPO         & 47.5 & 56.7 & 64.6 & 71.3 & 76.3 & 79.2 \\
    \rowcolor{ourshighlight}
    \textbf{MASPO} & \best{53.2} & \best{66.2} & \best{70.0} & \best{75.6} & \best{79.9} & 82.4 \\
    
    \midrule
    \multicolumn{7}{c}{\textit{Backbone: DeepSeek-R1-Distill-Qwen-7B (AIME 2025)}} \\
    \midrule
    GRPO         & 37.4 & 44.4 & 50.8 & 55.2 & 58.1 & 60.5 \\
    Clip Higher  & 37.4 & 44.9 & 52.1 & 58.2 & 63.3 & 67.2 \\
    DAC 400      & 36.2 & 43.4 & 50.6 & 56.3 & 59.6 & 61.7 \\
    Entropy Adv  & 34.4 & 41.0 & 46.4 & 50.3 & 54.0 & 58.1 \\
    BAPO         & 37.7 & 44.5 & 50.3 & 54.2 & 56.7 & 58.4 \\
    SAPO         & 35.3 & 41.4 & 47.4 & 51.6 & 55.3 & 58.0 \\
    \rowcolor{ourshighlight}
    \textbf{MASPO} & \best{42.9} & \best{50.7} & \best{57.3} & \best{63.4} & \best{68.9} & \best{73.2} \\
    
    \midrule
    \multicolumn{7}{c}{\textit{Backbone: DeepSeek-R1-Distill-Qwen-14B}} \\
    \midrule
    \multicolumn{7}{c}{\textit{Dataset: AIME 2024}} \\
    GRPO         & 60.0 & 69.3 & 76.0 & 79.8 & 81.9 & 82.9 \\
    \rowcolor{ourshighlight}
    \textbf{MASPO} & \best{63.5} & \best{72.4} & \best{77.7} & \best{81.4} & \best{84.4} & \best{86.1} \\
    \midrule
    \multicolumn{7}{c}{\textit{Dataset: AIME 2025}} \\
    GRPO         & 42.0 & 48.4 & 54.2 & 60.1 & 64.7 & 67.8 \\
    \rowcolor{ourshighlight}
    \textbf{MASPO} & \best{45.3} & \best{52.2} & \best{58.2} & \best{63.7} & \best{68.4} & \best{72.4} \\
    
    \bottomrule
    \end{tabular}
\end{table*}

\subsection{Ablation on Positive Risk Control}
\label{app:pos_risk_ablation}

We conducted a decoupling experiment on DeepSeek-R1-Distill-Qwen-1.5B with $\alpha=0.5$ and $\beta_\text{low}=0.03$, comparing $\beta_\text{high}=0$ against $\beta_\text{high}=0.03$. The training dynamics in Figure \ref{fig:pos_ablation} demonstrate that without risk control on positive samples ($\beta_\text{high}=0$), performance begins to degrade prematurely. Conversely, enabling positive risk control prevents early degeneration and promotes sustained exploration.

\subsection{Scalability and Detailed Performance}
\label{app:scalability_and_passk}

To further investigate the scalability of our approach, we applied MASPO to the larger DeepSeek-R1-Distill-Qwen-14B model. As shown in Table \ref{tab:14b_full_results}, MASPO demonstrates significant scalability, consistently outperforming the GRPO baseline across all benchmarks with a +2.8\% gain in average Avg@32. Furthermore, to provide a more fine-grained understanding of the generation quality, we report the detailed Pass@$k$ performance on AIME 2024 and AIME 2025 in Table \ref{tab:pass_k_detailed}. These results confirm that MASPO not only achieves higher accuracy (Pass@1) but also maintains a higher upper bound of correct solutions (Pass@32) compared to other strong baselines.

\end{document}